\documentclass{article}

\PassOptionsToPackage{numbers, compress}{natbib}

 \usepackage[final]{neurips_2026}

\usepackage[utf8]{inputenc} 
\usepackage[T1]{fontenc}    
\usepackage{hyperref}       
\usepackage{url}            
\usepackage{booktabs}       
\usepackage{amsfonts}       
\usepackage{nicefrac}       
\usepackage{microtype}      
\usepackage{xcolor}         
\usepackage{graphicx}
\usepackage{booktabs}
\usepackage{multirow}
\usepackage{makecell}
\usepackage{colortbl}
\usepackage[table]{xcolor}
\usepackage{comment}

\title{MechVerse: Evaluating Physical Motion Consistency in Video Generation Models}

\author{%
  \underline{Rahul Jain}$^{*1}$ \quad
  \underline{Mayank Patel}$^{*2}$ \quad
  \underline{Asim Unmesh}$^{1}$ \quad
  \underline{Karthik Ramani}$^{1,2}$ \\
  $^{1}$School of Electrical and Computer Engineering, Purdue University \\
  $^{2}$School of Mechanical Engineering, Purdue University \\
  \textit{$^{*}$Equal contribution}
}

\makeatletter

\providecommand{\@trackname}{Datasets and Benchmarks}

\makeatother
\begin{document}

\maketitle

\begin{abstract}

Text- and image-conditioned video generation models have achieved strong visual fidelity and temporal coherence, but they often fail to generate motion governed by kinematic and geometric constraints. In these settings, object parts must remain rigid, maintain contact or coupling with neighboring components, and transfer motion consistently across connected parts. These requirements are especially explicit in articulated mechanical assemblies, where motion is constrained by rigid-link geometry, contact/coupling relations, and transmission through kinematic chains. For example, gears rotate in coordination, cam--follower systems convert rotation into translation, and linkages propagate motion through connected rigid parts. A generated video may therefore appear plausible while violating the intended mechanism, such as rotating a part that should translate, deforming a rigid component, breaking coupling between parts, or failing to move downstream components. To evaluate this gap, We introduce MechVerse, a benchmark for mechanically consistent image-to-video generation. MechVerse contains 21,156 synthetic clips from 1,357 mechanical assemblies across 141 categories, organized into three tiers of increasing kinematic complexity: independent articulation, pairwise coupling, and densely coupled multi-part mechanisms. Each clip is paired with a structured prompt describing part identities, stationary supports, moving components, motion primitives, direction, speed/extent, and inter-part dependencies. We evaluate proprietary, open-source, and fine-tuned image-to-video models using standard video metrics, instruction-following scores, and human judgments of motion correctness and kinematic coupling. Results show that current models can preserve appearance and smoothness while failing to generate mechanically admissible motion, with errors increasing as coupling complexity grows. MechVerse provides a benchmark for measuring and improving mechanism-aware video generation from image and language inputs. The project page is: \href{https://mechverse.pages.dev/}{\textcolor{magenta}{https://mechverse.pages.dev/}}
\end{abstract}

\section{Introduction}\label{sec:intro}

\begin{figure*}[t]
\centering
\includegraphics[width=\linewidth]{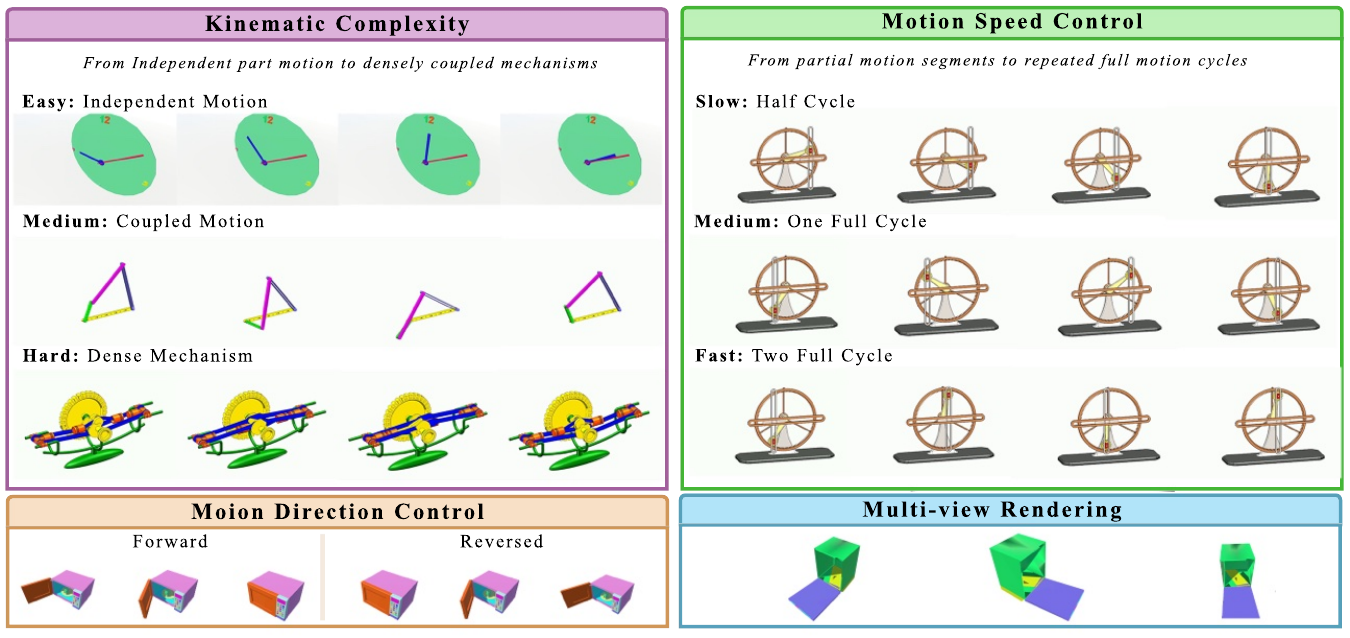}
\caption{
Overview of MechVerse and its structured motion variation design. \textbf{Top-left:} MechBench spans three levels of kinematic complexity, ranging from independent articulated motion (\textit{Easy}) to coupled mechanisms (\textit{Medium}) and densely interacting multi-part systems (\textit{Hard}). \textbf{Top-right:} Motion speed is systematically varied across partial, single-cycle, and repeated-cycle motion trajectories (\textit{Slow}, \textit{Medium}, \textit{Fast}). \textbf{Bottom-left:} Each motion sequence is paired with forward and reversed temporal directions, enabling semantically distinct motions such as opening/closing or clockwise/counter-clockwise rotation. \textbf{Bottom-right:} Assemblies are rendered from multiple camera viewpoints to provide viewpoint diversity while preserving identical underlying mechanical motion.
}
\label{fig:mechbench_teaser}
\end{figure*}

Recent image-to-video generation models can animate a single image from a text prompt with high visual fidelity and temporal smoothness~\citep{blattmann2023stable, yang2024cogvideox, kong2024hunyuanvideo, wan2025wan, ma2025step}. This progress has expanded the range of scenes and object motions that can be synthesized from simple image-language inputs. However, existing benchmarks mostly focus on natural scenes, human actions, broad object motion, or general visual quality~\citep{liu2024evalcrafter, liu2023fetv, huang2024vbench}. These benchmarks are useful for measuring perceptual quality, temporal coherence, and text-video alignment. But they do not test whether generated motion follows explicit part-level mechanical constraints. 

Mechanical assemblies expose this limitation. Their motion is governed by structured dependencies between parts. When one component moves, other components often need to move through joints, contact, or linkages. If these dependencies are violated, the video may still look smooth but become mechanically incorrect. This raises a key question: can current image-to-video models preserve part-level geometric and kinematic constraints? We study this question and find that existing video generation benchmarks have three limitations:

\begin{itemize}
    \item \textbf{Limited coverage of articulated mechanisms.}
    Existing benchmarks cover diverse scenes, actions, and object motions~\citep{liu2024evalcrafter, liu2023fetv, huang2024vbench}. However, they do not explicitly study mechanical assemblies, where motion is governed by coupled part-level constraints and rich geometric structure.

    \item \textbf{Insufficient control over motion dependencies.}
    Many benchmarks evaluate general motion quality or prompt alignment, but do not control whether motion is independent, pairwise coupled, or propagated through densely connected multi-part mechanisms. This makes it difficult to analyze how model performance changes with kinematic complexity.

    \item \textbf{Lack of structured and fine-grained motion specifications.}
    Existing prompts often describe motion at a semantic level, such as an object moving, rotating, or opening \citep{unmesh2023interacting, ji2020action}. They usually do not specify stationary supports, moving components, motion primitives, direction, speed or extent, and inter-part dependencies. Without such structure, it is difficult to test whether a generated video follows the intended mechanism.
\end{itemize}

Image-to-video generation is being applied to procedural and instructional 
content~\citep{miech2019howto100m,benshabat2021ikea}, world models and 
synthetic interaction data for embodied 
agents~\citep{yang2024unisim,xiang2020sapien}, and CAD-based design and 
digital-twin visualization, where the usefulness of a generated clip 
depends on whether parts move under the correct kinematic constraints 
rather than only on appearance.
Mechanically constrained motion is also a probe for the broader claim 
that video generation models implicitly learn world 
dynamics~\citep{openai2024sora,ha2018world}, since existing physics-oriented 
evaluations focus on rigid-body, fluid, and collision 
behavior~\citep{bansal2024videophy,meng2024phygenbench,motamed2025physicsiq} 
and rarely consider articulated mechanisms whose constraints are 
deterministic and propagate globally across coupled parts.
Varying dependency complexity from independent parts to densely coupled 
mechanisms further turns kinematic structure into a controlled axis of 
compositional generalization, of the kind where diagnostic 
benchmarks~\citep{johnson2017clevr,bear2021physion} have historically 
driven progress that aggregate video-quality 
scores~\citep{huang2024vbench,liu2024evalcrafter} cannot expose.
Closing this gap therefore matters application reliability, 
for scientific evaluation of generative world models, and for benchmark 
methodology.

To study this problem, we introduce \textbf{MechVerse}, a synthetic video benchmark for mechanically consistent image-to-video generation. Large-scale web video corpora provide broad appearance and motion priors, but offer little control over explicit part-level dependencies. MechVerse includes  controlled examples where the moving parts, stationary supports, motion primitives, and inter-part dependencies are known. MechVerse is constructed from two complementary sources: the PartNet-Mobility dataset~\citep{xiang2020sapien}, which provides kinematically annotated articulated objects spanning 46 categories, and a curated library of CAD mechanical assemblies covering linkages, cam-and-follower systems, gear trains, and complex multi-part mechanisms. All assemblies are rendered in Unity using a frame-accurate animation pipeline that drives joint motion via normalized time stepping, enabling precise control over motion coverage and reproducibility. The resulting dataset contains over 21156 videos from 1,357 assemblies clips organized into three tiers of increasing kinematic complexity: \textit{Easy} (single- or dual-part articulation), \textit{Medium} (two-part coupled mechanisms with 3--8 parts), and \textit{Hard} (strongly coupled multi-part assemblies with 10--50 parts) as shown in \autoref{fig:mechbench_teaser}. Each clip is systematically varied along three axes: motion speed (slow, mid, fast), camera viewpoint (canonical and mirrored), and motion direction (forward and reversed), to provide comprehensive coverage of the inference conditions under which a generative model must reason about mechanical structure. Each clip is paired with an input image and a structured text prompt describing part identities, stationary and moving components, motion type, direction, speed or extent, and inter-part dependencies. This setup lets us evaluate whether generated videos preserve the specified motion of individual parts as well as the dependencies between them. This setup provides a setting for studying whether image-to-video models can generate mechanically constrained motion.

Using MechVerse, we conduct the first large-scale benchmark of state-of-the-art video generation models on mechanically-consistent video synthesis. We evaluate 14 models that includes both open-source models, including DynamiCrafter~\citep{xing2024dynamicrafter}, ConsistI2V \cite{ren2024consisti2v}, and CogVideoX~\citep{yang2024cogvideox}, and closed-source systems such as KlingV3, Wan 2.7 and Happy Horse. We use standard video metrics, instruction-following scores, and human judgments of motion correctness and coupling \cite{huang2024vbench, li2025worldmodelbench}. Our analysis reveals three key findings: (i) perceptual video quality is a weak proxy for mechanical correctness; (ii) models often confuse motion primitives, fail to move driven components, or break coupling between interacting parts; and (iii) errors increase with kinematic complexity, especially for densely coupled multi-part mechanisms. Overall, MechVerse shows that current models can produce smooth and visually plausible videos, but do not yet reliably capture structured inter-part motion from image and language inputs.

\section{Related Work}
\label{sec:related}

\noindent\textbf{Video Generation.} Early text-to-video models extended image diffusion models with temporal modules~\citep{blattmann2023stable, guo2023animatediff, chen2024videocrafter2, wang2023modelscope}, while Diffusion Transformer-based systems~\citep{peebles2023scalable} enabled higher-fidelity generation in models such as CogVideoX~\citep{yang2024cogvideox}, HunyuanVideo~\citep{kong2024hunyuanvideo}, Wan2.1~\citep{wan2025wan}, Step-Video-T2V~\citep{ma2025step}, and Sora~\citep{openai_sora_2025_web}. Physical consistency in generation has been studied through evaluation benchmarks~\citep{bansal2024videophy, meng2024towards}, training-time annotation~\citep{wang2025wisa}, and prompt refinement~\citep{xue2025phyt2v}, while animation methods~\citep{le2023differentiable, montanaro2024motioncraft, liu2024physgen} animate scenes under simplified physical assumptions. These works address global physical realism but not structured kinematic coupling between mechanically interacting parts, which is the focus of MechVerse.

\noindent\textbf{Articulated Object Datasets.} PartNet-Mobility~\citep{xiang2020sapien} provides kinematic joint annotations for over 2,000 objects and is the most widely used articulated motion resource. Shape2Motion~\citep{wang2019shape2motion}, RPM-Net~\citep{yan2020rpm}, ACD~\citep{iliash2024s2o}, and GAPartNet~\citep{geng2023gapartnet} extend coverage to joint discovery, motion fields, and functional part semantics. \citet{patel2025dynamo} introduced a benchmark of 693 synthetic CAD gear assemblies for coupled motion prediction from point clouds. In the video domain, HA-ViD~\citep{zheng2023ha} and IKEA Video Manuals~\citep{liu2024ikea} provide assembly recordings with temporal annotations. All of these resources target 3D motion estimation or action recognition rather than 2D generative video evaluation, and none provides the structured prompt-paired clips with controlled kinematic complexity that MechVerse offers.

\section{MechVerse Dataset}
\label{sec:dataset}
MechVerse comprises 21,156 video clips derived from 1,357 unique mechanical assemblies spanning 141 categories, constructed from two complementary sources: 904 assemblies from PartNet-Mobility~\citep{xiang2020sapien}, covering everyday articulated objects with independent part motion assigned to the \textit{Easy} tier, and 453 CAD mechanical assemblies curated specifically for this dataset covering linkages, cam-and-follower systems, engine pistons, and complex multi-part mechanisms, assigned to the \textit{Medium} and \textit{Hard} tiers. The following subsections describe the complexity structure, clip variation axes, prompt design, annotation pipeline, and train/test split in detail.

\begin{figure}[t]
\centering
\includegraphics[width=\linewidth]{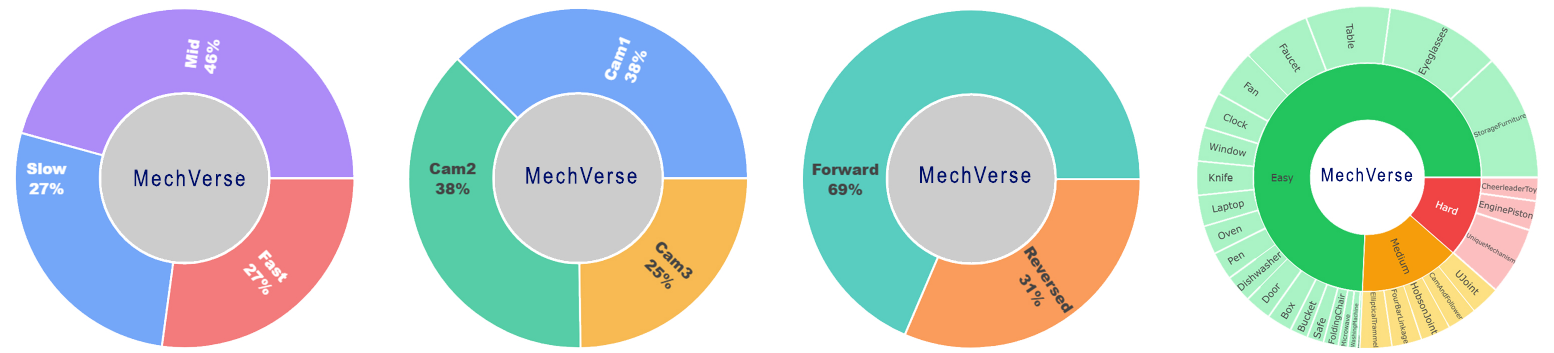}
\caption{MechVerse dataset statistics. \textit{Top row:} Distribution of clips by speed variant (Slow 27\%, Mid 46\%, Fast 27\%), camera viewpoint (Cam1 25\%, Cam2 38\%, Cam3 37\%), motion direction (Forward 69\%, Reversed 31\%), and category hierarchy by complexity tier.}
\label{fig:dataset_stats}
\end{figure}

\subsection{Complexity Tiers}
A single aggregate evaluation score cannot reveal \emph{where} a generative model fails, whether it struggles with any motion at all, or only when motion must propagate across interacting parts. To enable this diagnostic analysis, MechVerse organizes assemblies into three tiers of kinematic complexity, reflecting a fundamental distinction between independent and coupled part motion.

\textit{Easy} assemblies (15,720 clips) contain one or two parts where motion is independent: a door swinging, a clock hand rotating, or a drawer sliding, consistent with the motion structure of prior articulated object datasets. \textit{Medium} assemblies (2,412 clips) contain 3-8 parts where motion is kinematically coupled: a rotating input drives other components through linkages or contact, as in four-bar linkages, cam-and-follower mechanisms, and Hobson joints. \textit{Hard} assemblies (3,024 clips) contain 10 or more parts, where dense kinematic coupling and frequent self-occlusion in 2D projection make motion understanding significantly more challenging. This tiered structure enables progressive evaluation of generative model behavior as interaction complexity increases, and allows failures to be localized to specific regimes of motion complexity rather than averaged away.

\subsection{Clip Variation}
To prevent models from exploiting spurious correlations between visual appearance and motion behavior, each assembly is rendered under systematic variation along three axes: motion speed, camera viewpoint, and motion direction. Speed variants are defined by the fraction of the input joint range covered: \textit{Slow} covers the first half of the joint range, \textit{Mid} covers the full range, and \textit{Fast} repeats the full range twice. For CAD assemblies, these correspond to 180$^\circ$, 360$^\circ$, and 720$^\circ$ of input rotation respectively. All clips are rendered at 16 FPS for a duration of 2 seconds, yielding 32 frames per clip. PartNet-Mobility assemblies are rendered from three camera viewpoints (Cam1, Cam2, Cam3) placed at fixed positions in the Unity virtual environment, while CAD assemblies are rendered from two viewpoints. Reversed clips are generated by reversing the frame order of forward clips, yielding semantically distinct motion directions (e.g., clockwise vs.\ counter-clockwise, opening vs.\ closing, extending vs.\ retracting). The dataset contains 14,508 forward and 6,648 reversed clips.

\subsection{Prompt Structure}
\label{sec:prompt_structure}

\begin{figure}[b]
\centering
\includegraphics[width=\linewidth]{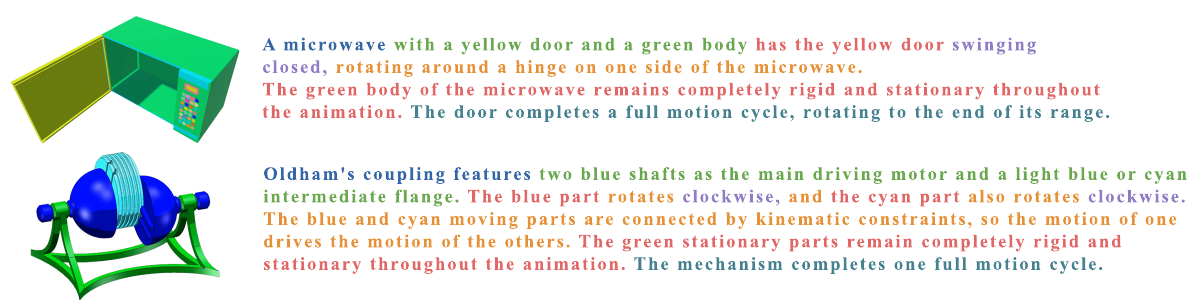}
\caption{Representative MechVerse prompt examples with color-coded components: \textcolor[RGB]{52,101,164}{assembly overview}, \textcolor[RGB]{106,168,79}{part identification}, \textcolor[RGB]{224,102,102}{moving/rigid classification}, \textcolor[RGB]{230,145,56}{motion type}, \textcolor[RGB]{142,124,195}{direction}, and \textcolor[RGB]{69,129,142}{speed description}. Top: Easy-tier example (microwave). Bottom: Hard-tier example (Oldham's coupling) where multiple parts are kinematically coupled.}
\label{fig:prompt_samples}
\end{figure}

For a video generation model to produce mechanically correct output, it must receive conditioning information that is precise, unambiguous, and complete. A vague prompt such as ``a box opening'' leaves the model free to hallucinate motion that looks plausible but violates the assembly's kinematic structure. To address this, each clip in MechVerse is paired with a structured natural language prompt composed of six semantically distinct components: \textcolor[RGB]{52,101,164}{\textbf{(1) assembly overview}}, \textcolor[RGB]{106,168,79}{\textbf{(2) part identification}} by flat matte color, \textcolor[RGB]{224,102,102}{\textbf{(3) moving/rigid classification}} per colored part, \textcolor[RGB]{230,145,56}{\textbf{(4) motion type}} (rotation, translation, rotation+translation, or planar), \textcolor[RGB]{142,124,195}{\textbf{(5) direction of motion}} (e.g., clockwise, closing, sliding left), and \textcolor[RGB]{69,129,142}{\textbf{(6) speed description}} varying across slow, mid, and fast variants. This fixed ordering ensures every prompt is grounded in expert mechanical knowledge and fully characterizes the clip's visual and kinematic content. Figure~\ref{fig:prompt_samples} illustrates two representative examples with each component color-coded accordingly.

\begin{figure*}[h]
\centering
\includegraphics[width=\linewidth]{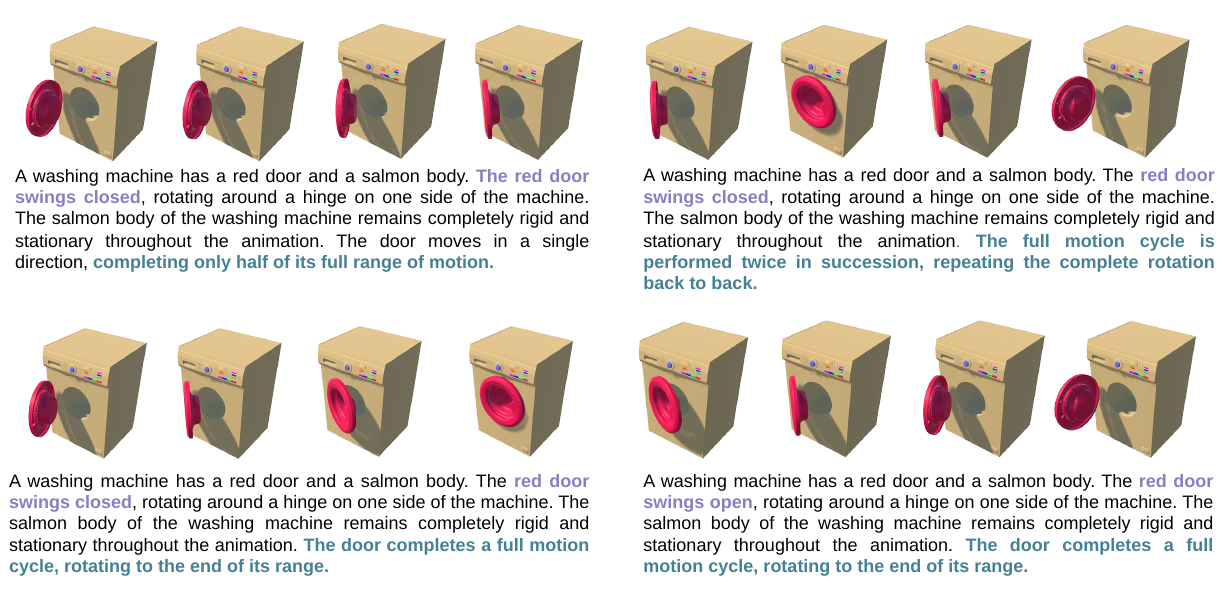}
\caption{Fine-grained motion control in MechVerse via prompt variation. All four rows share the same input image (a washing machine with a red door and a salmon body). Varying the \textcolor[RGB]{142,124,195}{\textbf{direction}} and \textcolor[RGB]{69,129,142}{\textbf{speed}} components of the prompt produces four semantically distinct ground-truth clips: \textit{Row 1}: \textcolor[RGB]{142,124,195}{closing}, \textcolor[RGB]{69,129,142}{slow} (half range); \textit{Row 2}: \textcolor[RGB]{142,124,195}{closing}, \textcolor[RGB]{69,129,142}{mid} (full cycle); \textit{Row 3}: \textcolor[RGB]{142,124,195}{opening}, \textcolor[RGB]{69,129,142}{mid} (full cycle); \textit{Row 4}: \textcolor[RGB]{142,124,195}{closing}, \textcolor[RGB]{69,129,142}{fast} (two full cycles).}
\label{fig:variation_samples}
\end{figure*}

\subsection{Fine-Grained Motion Control via Prompt Variation}
\label{sec:prompt_variation}
A key design principle of MechVerse is that the same input image can condition generation of multiple semantically distinct animations by varying only the prompt. Because speed, direction, and motion extent are explicitly encoded as separate components of each prompt, the dataset enables fine-grained control over motion behavior without any change to the visual input. Figure~\ref{fig:variation_samples} illustrates this with four clips derived from the same washing machine assembly: varying the speed description produces animations where the door completes half its range (slow), one full cycle (mid), or two full cycles (fast), while varying the direction keyword switches the motion from closing to opening. This structured variation is present across all 1,357 assemblies in MechVerse and constitutes a systematic evaluation axis that existing articulated object datasets do not provide.

\subsection{Prompt Annotation}
Automating prompt generation via MLLMs failed systematically on Medium and Hard assemblies,  coupled motion types were consistently misclassified, for example a rotating cam labeled as linear motion. Full details and failure examples are provided in Appendix~\ref{app:mllm_annotation}. We therefore developed an expert annotation pipeline in which two mechanical engineering annotators labeled each clip using a custom web tool (see Appendix~\ref{app:annotation_pipeline}), recording part colors, moving/stationary classification, motion type, and direction keywords. Structured fields were then assembled into a draft prompt and reformatted into fluent prose using GPT-4o mini under a strict system prompt that prohibited hallucination and omission; the LLM served only as a formatter. Final prompts were verified in a second review application (see Appendix~\ref{app:annotation_pipeline}). MechVerse releases video clips, prompts, and underlying 3D assets (OBJ files) for all assemblies.

\subsection{Train/Test Split}
A common failure mode in dataset design is test contamination through visual similarity, if a model has seen assemblies of the same category during training, strong performance on the test set may reflect memorization rather than generalization. To prevent this, MechVerse uses a category-level holdout split in which no assembly category present in the test set appears in the training set.

The dataset is divided into 18,672 training clips and 1,004 test clips. The test set covers 8 held-out categories (Bucket, CamAndFollower, FoldingChair, Microwave, Refrigerator, Safe, UniqueMechanism, WashingMachine) spanning all three complexity tiers, with 476 Easy, 294 Medium, and 234 Hard clips. To ensure that test evaluation measures motion understanding rather than viewpoint generalization, all test clips are rendered from a single standardized camera viewpoint (Cam1). An additional 1,480 clips corresponding to alternative camera views of test assemblies are excluded from both splits.

\section{Experiments}

\textbf{Evaluation Setup.} We use MechVerse to evaluate whether image-to-video models can generate motion that is both visually coherent and consistent with the geometric and kinematic constraints of mechanical assemblies. The evaluation is performed on a test set that contains assemblies with varying levels of complexity (easy, medium, and hard), ensuring coverage of diverse interaction scenarios. Each model takes an input image and a textual description that includes object-level motion and interaction constraints and generates a video. We evaluate the generated outputs directly using default inference settings for each model, without additional modifications.No additional post-processing or normalization is applied.

\textbf{Evaluation Models.}
We evaluate a broad set of recent image-to-video systems, including proprietary models, open-source models, and fine-tuned variants. The proprietary models include Wan~2.7~\cite{wan27web},  Kling~3~\cite{kling3web}, and Happy Horse~1.0~\cite{happyhorse10web}. The open-source models include Wan~2.2~\cite{wan2025wan}, CogVideoX~\cite{yang2024cogvideox}, CogVideoX~1.5~\cite{yang2024cogvideox}, HunyuanVideo~\cite{kong2024hunyuanvideo}, VideoCrafter~\cite{chen2024videocrafter2}, DynamiCrafter~\cite{xing2024dynamicrafter}, ConsistI2V~\cite{ren2024consisti2v}, LTX-Video~\cite{hacohen2024ltxvideo}, and LTX-2~\cite{hacohen2026ltx2}. We also evaluate fine-tuned variants of Wan~2.2 and CogVideoX trained on the MechVerse training split. These models represent a diverse set of recent approaches for image-to-video generation, spanning diffusion-based video generation, transformer-based video diffusion, and image-conditioned generation paradigms~\cite{blattmann2023stable,chen2024videocrafter2,girdhar2024factorizing,wan2025wan,kong2024hunyuanvideo}. All models are evaluated under a consistent setup using the same input image and prompt pairs.

\textbf{Metrics.}
We use VBench-I2V\cite{huang2024vbench} to measure 
standard image-to-video quality, including temporal consistency, motion smoothness, visual fidelity, and consistency with the input image. This covers metrics such assubject consistency, background consistency, motion smoothness, temporal flickering, dynamic degree, aesthetic quality, imaging quality, and I2V subject fidelity.  We further include WorldModelBench \cite{li2025worldmodelbench} metrics to evaluate
whether the generated videos follow the instruction, remain physically plausible, and satisfy basic common-sense constraint (frame-wise and temporal quality). We exclude the fluid dynamics sub-category from the physics score as our dataset contains only rigid-body mechanical assemblies. We adopt this benchmark as our prompts describe mechanical assembly animations where models must follow explicit motion instructions and respect physical laws and constraints.

\textbf{Training Details.}
We fine-tune both models using LoRA with rank 32 for 2 epochs on the full training set. We use a batch size of 1 for all fine-tuning experiments. All experiments are conducted on an NVIDIA 4-way GH200 GPU cluster with an aarch64 system architecture.

\begin{table*}[t]
\caption{Evaluation results on \textbf{MechVerse}. VBench I2V metrics (scores in \%): \textbf{Subj.}~=~Subject Consistency; \textbf{BG Cons.}~=~Background Consistency; \textbf{Motion}~=~Motion Smoothness; \textbf{T.Flick.}~=~Temporal Flickering \textbf{Dyn.}~=~Dynamic Degree; \textbf{Aesth.}~=~Aesthetic Quality; \textbf{Imaging}~=~Imaging Quality; 
\textbf{I2V Subj.}~=~I2V Subject fidelity. WorldModelBench metrics: \textbf{Instr.}~=~Instruction-following score; \textbf{Physics}~=~Physical law adherence; \textbf{Common}~=~Common-sense reasoning; \textbf{Sum}~=~aggregate of the three. \colorbox[HTML]{FFD700}{\strut Gold}~/~\colorbox[HTML]{C0D8F0}{\strut Blue}~/~\colorbox[HTML]{D5F5E3}{\strut Green} = 1st~/~2nd~/~3rd best per column.}
\label{tab:mechbench}
\centering
\resizebox{\textwidth}{!}{%
\begin{tabular}{l|cccccccc|cccc}
\toprule
 & \multicolumn{8}{c|}{VBench I2V} & \multicolumn{4}{c}{WorldModelBench} \\
\cmidrule(lr){2-9} \cmidrule(lr){10-13}
Model & Subj. & BG Cons. & Motion & T.Flick. & Dyn. & Aesth. & Imaging & I2V Subj. & Instr. & Physics & Common & Sum \\
\midrule
\multicolumn{13}{c}{\textit{Proprietary Models}} \\
\midrule
Wan2.7 & 93.65 & \cellcolor[HTML]{C0D8F0}\textbf{96.07} & 99.22 & 99.20 & 34.36 & 51.04 & \cellcolor[HTML]{C0D8F0}\textbf{64.09} & 96.98 & \cellcolor[HTML]{C0D8F0}\textbf{2.00} & 3.70 & \cellcolor[HTML]{D5F5E3}1.90 & \cellcolor[HTML]{C0D8F0}\textbf{7.59} \\
Happy Horse & \cellcolor[HTML]{C0D8F0}\textbf{94.95} & \cellcolor[HTML]{D5F5E3}96.03 & 99.43 & \cellcolor[HTML]{FFD700}\textbf{99.41} & 18.73 & 50.83 & 58.68 & \cellcolor[HTML]{C0D8F0}\textbf{98.41} & \cellcolor[HTML]{D5F5E3}1.95 & 3.70 & \cellcolor[HTML]{C0D8F0}\textbf{1.91} & \cellcolor[HTML]{D5F5E3}7.56 \\
Kling3 & \cellcolor[HTML]{D5F5E3}94.91 & 95.28 & \cellcolor[HTML]{D5F5E3}99.48 & \cellcolor[HTML]{D5F5E3}99.29 & 37.55 & \cellcolor[HTML]{C0D8F0}\textbf{51.65} & 56.07 & \cellcolor[HTML]{D5F5E3}98.15 & \cellcolor[HTML]{FFD700}\textbf{2.02} & \cellcolor[HTML]{D5F5E3}3.71 & \cellcolor[HTML]{FFD700}\textbf{1.93} & \cellcolor[HTML]{FFD700}\textbf{7.66} \\
\midrule
\multicolumn{13}{c}{\textit{Open-Source Models}} \\
\midrule
CogVideoX-1.0 (5B) & 89.78 & 94.29 & 98.83 & 98.60 & 29.58 & 48.88 & 54.37 & 93.70 & 1.33 & 3.13 & 1.32 & 5.79 \\
CogVideoX-1.5 (5B) & 87.37 & 92.63 & 99.04 & 98.72 & \cellcolor[HTML]{D5F5E3}50.20 & 45.09 & 57.28 & 93.93 & 1.34 & 3.17 & 1.16 & 5.68 \\
DynamiCrafter & 93.43 & 95.15 & 99.31 & 98.66 & 25.90 & 47.57 & 57.98 & 95.25 & 1.21 & \cellcolor[HTML]{C0D8F0}\textbf{3.75} & 1.71 & 6.68 \\
HunyuanVideo-1.5 (8.3B)& \cellcolor[HTML]{FFD700}\textbf{95.61} & \cellcolor[HTML]{FFD700}\textbf{96.78} & \cellcolor[HTML]{C0D8F0}\textbf{99.49} & \cellcolor[HTML]{C0D8F0}\textbf{99.35} & 12.85 & \cellcolor[HTML]{FFD700}\textbf{52.41} & 56.89 & \cellcolor[HTML]{FFD700}\textbf{98.48} & 1.88 & \cellcolor[HTML]{FFD700}\textbf{3.76} & \cellcolor[HTML]{FFD700}\textbf{1.93} & \cellcolor[HTML]{D5F5E3}7.56 \\
LTX-2 (22B)& 90.66 & 93.80 & 99.18 & 99.01 & 44.52 & \cellcolor[HTML]{D5F5E3}51.08 & 55.20 & 94.89 & 1.67 & 3.27 & 1.37 & 6.31 \\
LTX-Video (13B)& 93.04 & 95.63 & \cellcolor[HTML]{FFD700}\textbf{99.51} & 99.16 & 50.00 & 50.96 & \cellcolor[HTML]{D5F5E3}60.26 & 95.63 & 1.44 & 3.56 & 1.57 & 6.57 \\
VideoCrafter & 93.30 & 94.91 & 97.53 & 96.34 & \cellcolor[HTML]{FFD700}\textbf{65.24} & 47.59 & \cellcolor[HTML]{FFD700}\textbf{70.00} & 88.85 & 1.24 & 3.64 & 1.71 & 6.58 \\
Wan2.2 (5B)& 89.64 & 92.57 & 98.45 & 98.10 & 44.02 & 47.89 & 58.59 & 95.13 & 1.63 & 3.36 & 1.57 & 6.56 \\
ConsistI2V & 89.85 & 92.52 & 98.80 & 98.06 & 19.96 & 42.45 & 58.77 & 91.60 & 1.12 & 3.47 & 1.06 & 5.65 \\
\midrule
\multicolumn{13}{c}{\textit{Fine-tuned Models}} \\
\midrule
CogVideoX-1.0 (FT) (5B)& 93.13 & 94.90 & 99.18 & 99.13 & 14.54 & 47.72 & 50.04 & 94.59 & 1.47 & 3.54 & 1.64 & 6.64 \\
Wan2.2 (FT) (5B)& 90.44 & 93.36 & 98.54 & 98.13 & \cellcolor[HTML]{C0D8F0}\textbf{51.89} & 49.50 & 55.62 & 93.71 & 1.43 & 3.31 & 1.41 & 6.15 \\
\bottomrule
\end{tabular}%
}
\end{table*}

\section{Results}
\label{sec:results}

\subsection{Quantitative Results}
We report quantitative results in \autoref{tab:mechbench}. In general, all models obtain strong VBench-I2V scores, with most models achieving motion smoothness above 98 and temporal flickering close to 99. This suggests that current models can generate visually stable videos. However, they still struggle on WorldModelBench, where the task requires following motion instructions and preserving physically plausible motion. We find that Kling3 is the strongest closed-source model with the highest aggregate score of 7.66, followed by Wan2.7 and Horse with 7.59 and 7.56. Closed-source models are also stronger on instruction following, while open-source models perform noticeably worse on this metric.

Interestingly, open-source models remain competitive on perceptual video quality. HunyuanVideo-1.5 achieves the best scores on several VBench metrics, and VideoCrafter achieves the highest imaging quality. These findings show that open-source models can match or exceed closed-source models on standard perceptual metrics, but are less reliable when evaluation focuses on instruction following and physical plausibility.

We further break down results along three axes of our dataset: assembly complexity, motion speed, and direction. \textbf{Assembly complexity.} As shown in section~\ref{app:complexity}, VBench metrics slightly drop from Easy to Hard, but the differences are small. Motion smoothness and temporal flickering remain high across complexity levels, so VBench does not strongly separate Easy, Medium, and Hard cases.On WorldModelBench, however, we observe the opposite trend where scores actually go up with complexity. Scores often increase with complexity for example HunyuanVideo-1.5 goes from 7.26 on Easy to 7.67 on Medium and 8.03 on Hard, Happy Horse from 7.14 to 7.80 and 8.12, and DynamiCrafter from 6.34 to 6.86 and 7.12. Overall, complexity exposes a mismatch between automatic video-quality metrics and mechanism-level correctness. \textbf{Motion type.}  As shown in section~\ref{app:speed}, we further split the results by forward and reversed motion directions. In general, the gap is small, but reversed motions perform slightly better for most models in WorldModelBench. For example, HunyuanVideo-1.5 increases from 7.44 on forward motion to 7.75 on reversed motion. However, we see the opposite trend in VBench metrics. Across all models, reversed clips consistently perform lower on subject consistency, background consistency, and I2V subject fidelity compared to their forward motion. \textbf{Effect of motion speed.}
section~\ref{app:speed} shows that speed separates models the least in WorldModelBench, with most scores changing by less than 0.2 across slow, medium, and fast clips. Proprietary models improve slightly on fast clips, while most open-source models remain flat or slightly degrade. For VBench metric, we observe opposite trend where Faster videos are harder for all models. Overall Finetuning results are mixed. \textbf{CogVideoX} improves 
on nearly every metric, with its WorldModelBench aggregate rising from  \textbf{5.79 to 6.64}. \textbf{Wan2.2}, in contrast, gains in perceptual 
quality and dynamic degree but loses ground on physical-law adherence and common-sense reasoning, with its aggregate dropping from \textbf{6.56 to 6.15}. 
Mechanism-aware fine-tuning therefore helps weaker base models  substantially, but for stronger models perceptual gains can come at the cost of physical reasoning.

\begin{figure*}[h]
    \centering
    \includegraphics[width=\linewidth]{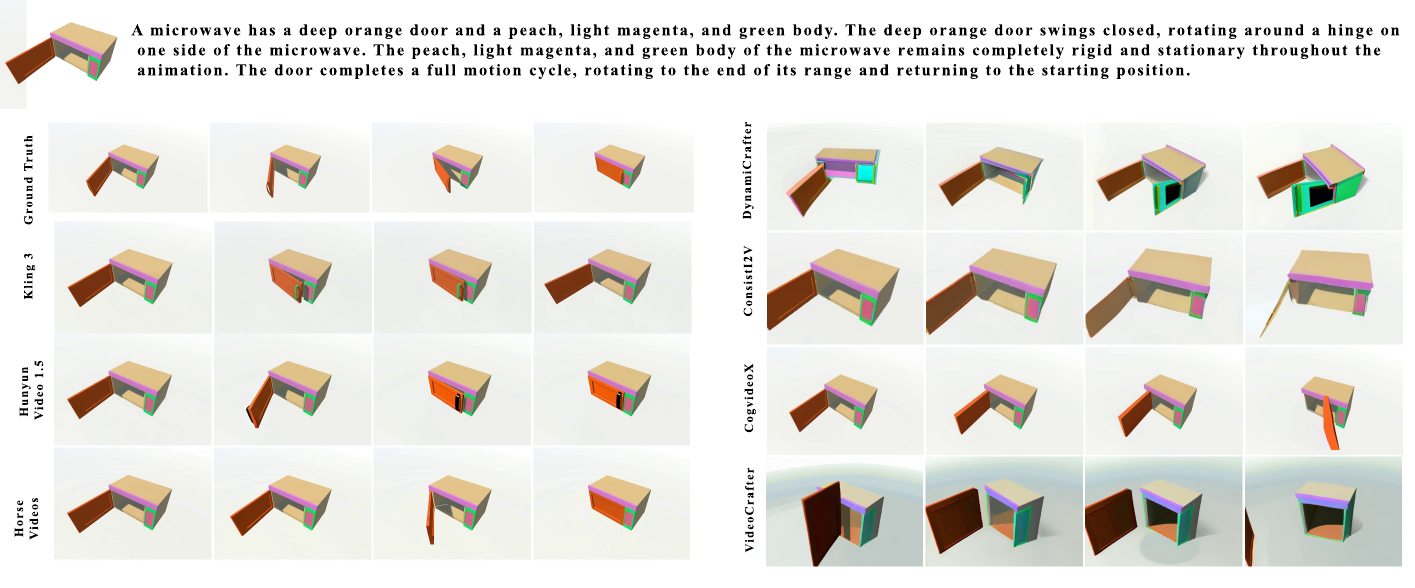}
    \caption{Qualitative comparison of generated videos on a Easy-tier MechVerse assembly.}
    \label{fig:qualitative}
\end{figure*}

\section{Analysis}

\subsection{Quantitative Results by Complexity Tier}
\label{app:complexity}

\begin{figure*}[h]
\centering
\includegraphics[width=\linewidth]{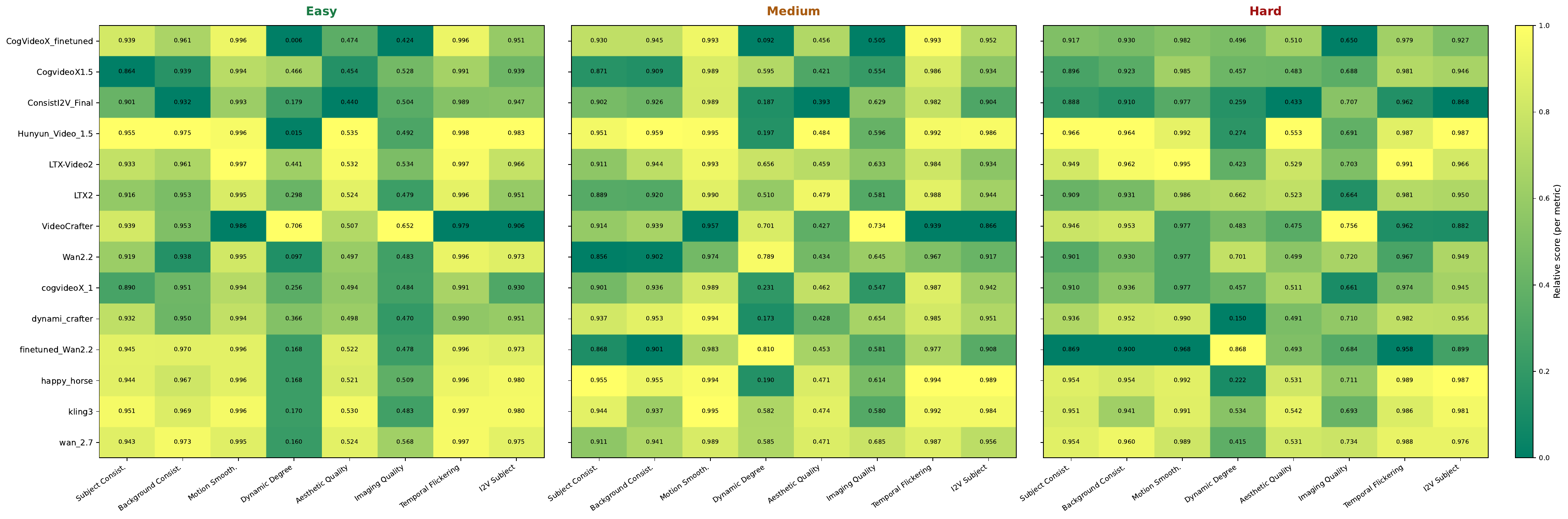}
\caption{VBench-I2V scores stratified by kinematic complexity tier (Easy / Medium / Hard). Perceptual metrics remain largely stable across tiers, indicating that VBench does not capture the degradation in mechanical correctness as coupling complexity increases.}
\label{fig:vbench_complexity}
\end{figure*}

\begin{figure*}[h]
\centering
\includegraphics[width=\linewidth]{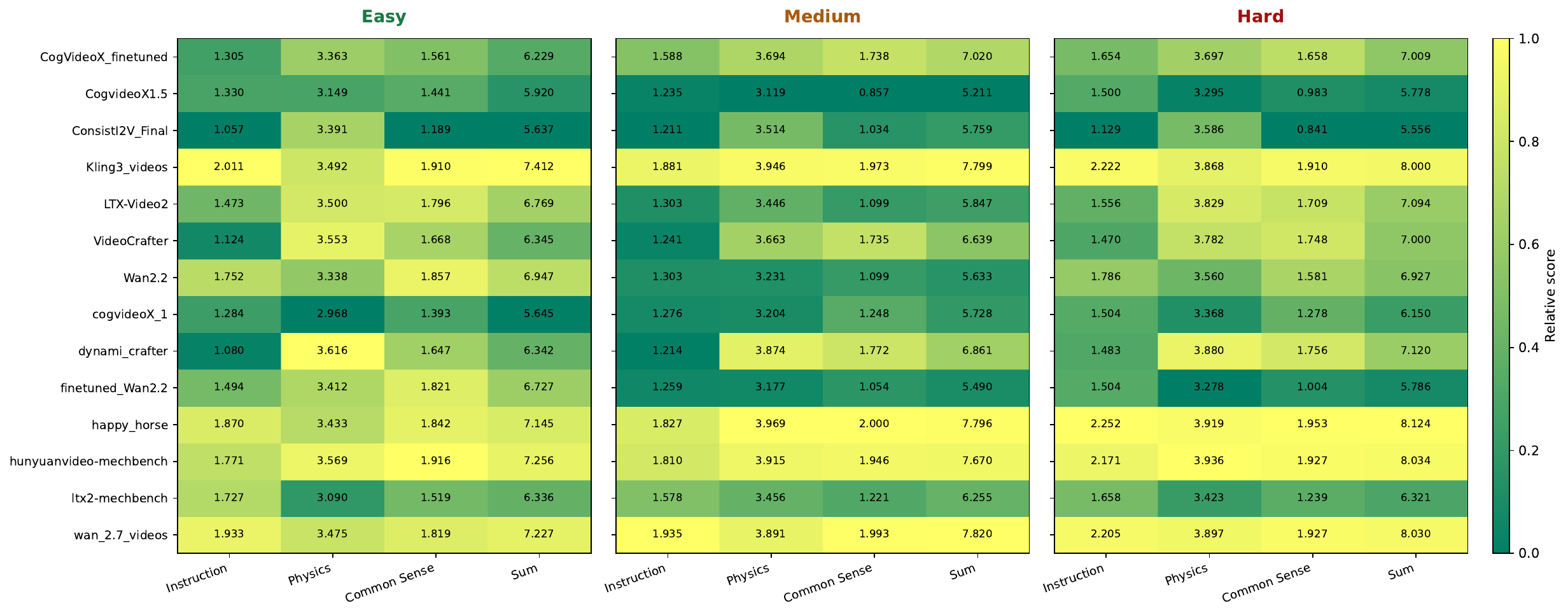}
\caption{WorldModelBench scores stratified by kinematic complexity tier (Easy / Medium / Hard). Aggregate scores increase with complexity, reflecting the tendency of MLLM judges to rate visually plausible but mechanically incorrect videos favourably on complex assemblies.}
\label{fig:wmb_complexity}
\end{figure*}

Figures~\ref{fig:vbench_complexity} and~\ref{fig:wmb_complexity} show VBench-I2V and WorldModelBench scores stratified by kinematic complexity tier. On VBench metrics, scores are largely stable across Easy, Medium, and Hard assemblies, motion smoothness and temporal flickering remain high throughout, confirming that perceptual quality does not degrade meaningfully with complexity. On WorldModelBench, we observe the opposite: aggregate scores tend to increase from Easy to Hard. For example, Kling3 goes from 7.41 on Easy to 7.80 on Medium and 8.00 on Hard, and Horse from 7.15 to 7.80 and 8.12. This counterintuitive trend reflects a known limitation of MLLM-based judges on complex assemblies: as coupling complexity increases, videos that appear globally plausible receive higher scores even when part-level kinematic constraints are violated. Human evaluation (Section~\ref{sec:results}) reveals the opposite trend, with motion correctness and kinematic coupling scores falling on Medium and Hard assemblies.

\subsection{Quantitative Results by Motion Speed and Direction}
\label{app:speed}

\begin{figure*}[h]
\centering
\includegraphics[width=\linewidth]{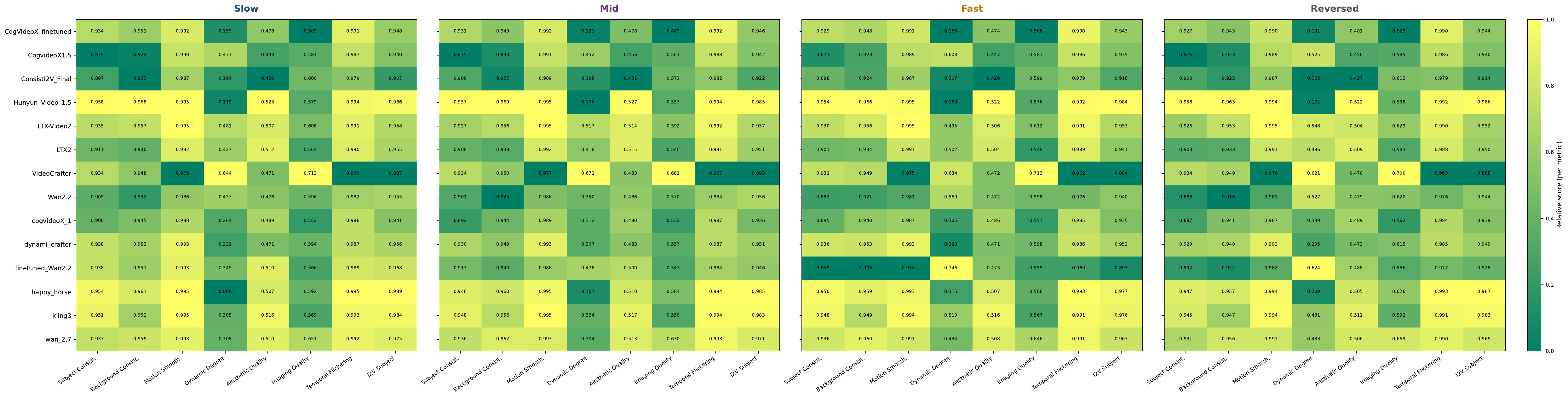}
\caption{VBench-I2V scores stratified by motion speed (Slow / Mid / Fast) and direction (Reversed). Faster clips degrade perceptual quality metrics across most models.}
\label{fig:vbench_speed}
\end{figure*}

\begin{figure*}[h]
\centering
\includegraphics[width=\linewidth]{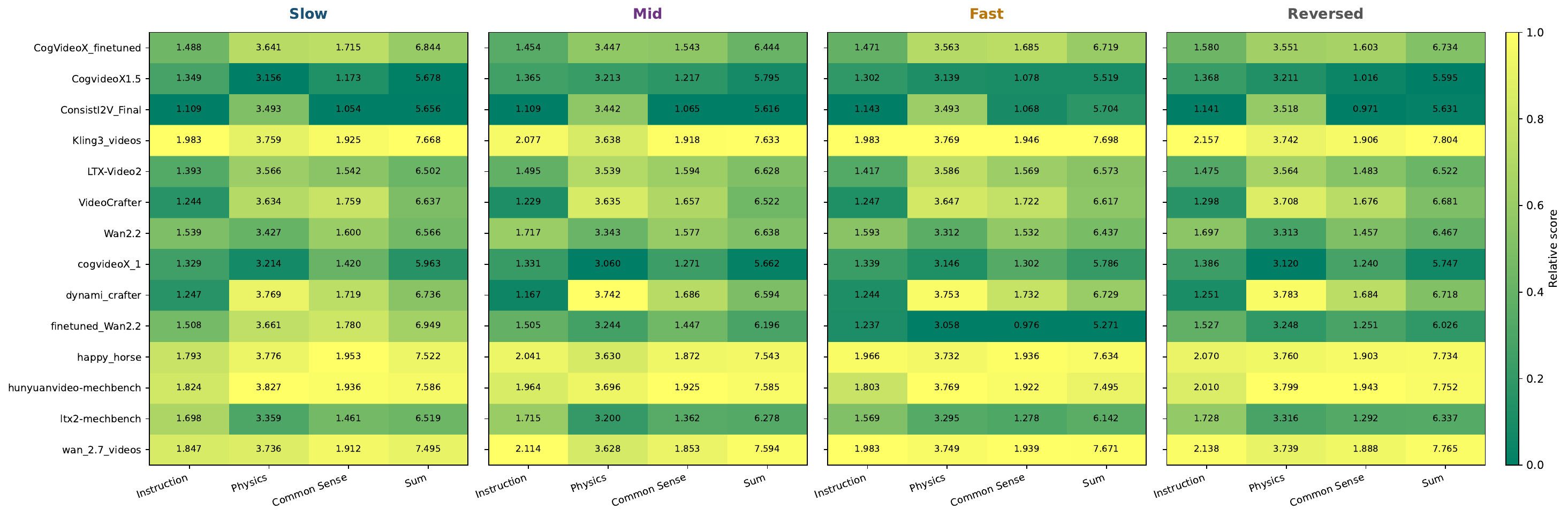}
\caption{WorldModelBench scores stratified by motion speed (Slow / Mid / Fast) and direction (Reversed). Speed has the smallest effect on WorldModelBench scores relative to complexity or direction.}
\label{fig:wmb_speed}
\end{figure*}

Figures~\ref{fig:vbench_speed} and~\ref{fig:wmb_speed} show scores split by motion speed (Slow / Mid / Fast) and direction (Reversed). On WorldModelBench, speed has the smallest effect of all axes, most models change by less than 0.2 across speed variants. Proprietary models improve slightly on fast clips while most open-source models remain flat or degrade marginally. On VBench, the opposite holds: faster clips are harder for all models, with dynamic degree and subject consistency both declining at higher speeds. For direction, reversed clips score slightly higher on WorldModelBench for most models (e.g., HunyuanVideo-1.5 goes from 7.59 on forward to 7.75 on reversed) but lower on VBench consistency metrics, suggesting that reversed motion is easier to rate as plausible but harder to render with visual fidelity.
\subsection{Human Evaluation}

\begin{figure*}[h]
    \centering
    \includegraphics[width=\linewidth]{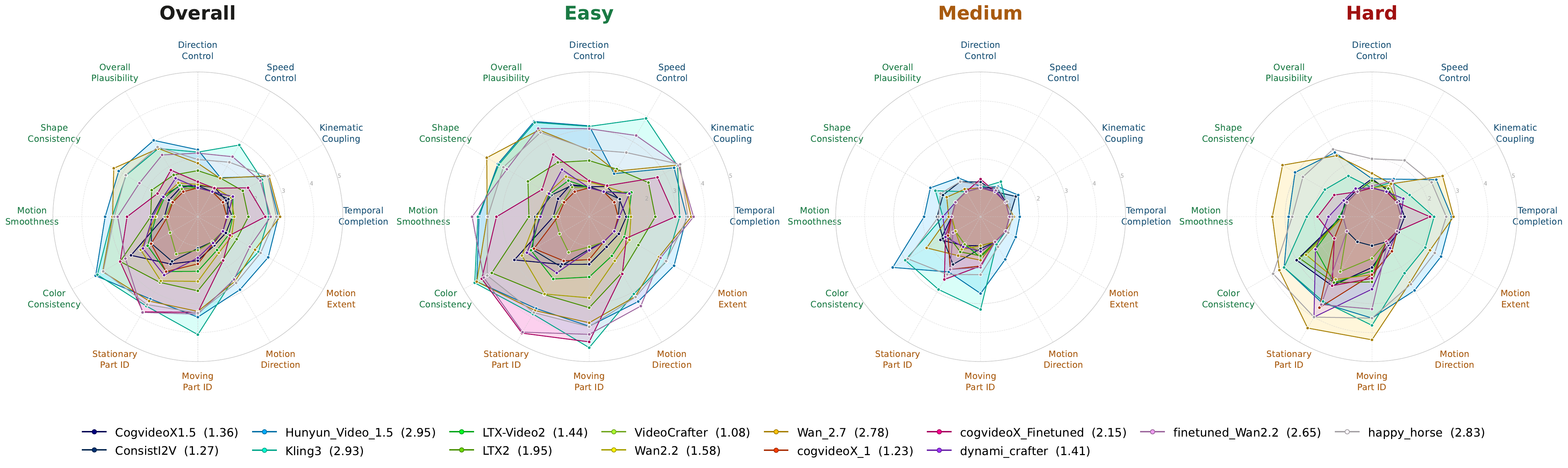}
    \caption{Human evaluation results across 14 models on 12 dimensions, grouped into three evaluation axes: \textcolor[RGB]{26,82,118}{\textbf{Motion Correctness}}, \textcolor[RGB]{26,122,68}{\textbf{Visual Realism}}, and \textcolor[RGB]{168,90,15}{\textbf{Prompt Adherence}}. }

    \label{fig:human_eval_radar}
\end{figure*}

We conduct human evaluation on 43 assemblies sampled from the test set, stratified by complexity level and motion category. Overall the total output is rated by six evaluators on 12 dimensions covering \textcolor[RGB]{26,82,118}{\textbf{motion correctness}}, \textcolor[RGB]{26,122,68}{\textbf{visual realism}}, and \textcolor[RGB]{168,90,15}{\textbf{prompt adherence}}, using a 1--5 Likert scale (Figure~\ref{fig:human_eval_radar}). Overall, human scores remain modest across models, showing that mechanism-aware video generation is still challenging. HunyuanVideo~1.5 achieves the highest overall score among the evaluated models (2.91), followed by Horse (2.89), Wan2.7 (2.78), and fine-tuned Wan2.2 (2.65). The largest gaps appear on motion-related axes such as correct part motion, direction following, temporal completion, and kinematic coupling. While several models preserve visual appearance, they often fail to generate physically consistent part motion. Performance also drops from Easy to Medium and Hard examples. This degradation is clearer in human evaluation than in VBench, where scores remain relatively consistent. We observe reliable inter-rater agreement, with Krippendorff's $\alpha$ of 0.72 for motion correctness, 0.63 for visual realism, and 0.77 for prompt adherence. Agreement is higher for prompt adherence and motion correctness than for visual realism. This suggest that evaluators can more consistently judge whether a video follows the instruction and preserves the intended mechanism. Together, these results show that human evaluation captures mechanism-level failures that are often missed by standard perceptual metrics.


We list the analysis of our results below:
\begin{itemize}
    \item \textbf{Automatic perceptual metrics are weak proxies for mechanism-aware generation.}
    Most models obtain high VBench-I2V scores, indicating strong appearance preservation and temporal stability. However, these metrics do not assess whether the generated motion satisfies the mechanical constraint in the prompt. As a result, visually smooth videos may still contain wrong part motion, deformation, reversed direction, or broken coupling.

    \item \textbf{MLLM judges struggle with fine-grained mechanical failures.}
    WorldModelBench provides better separation between models than VBench-I2V. However, its trends do not always align with human evaluation as assembly complexity increases. Many mechanical failures are local and subtle, such as incorrect motion of small parts, missing dependent motion, or motion that breaks over time.

    \item \textbf{Human evaluation reveals the main bottleneck.}
    Human scores show a clear gap between appearance-related and motion-related criteria. Models preserve color, shape, and overall plausibility more reliably than direction control, speed control, temporal completion, and kinematic coupling. This suggests that the main bottleneck is mechanism-level motion reasoning rather than visual fidelity.

    \item \textbf{Fine-tuning gives limited improvement.}
    Fine-tuned models improve over their zero-shot baselines on some WorldModelBench scores. This indicates that training on MechBench can improve mechanical prompt following. However, persistent errors in part motion, direction following, and coupling suggest that existing video models need more explicit representations of parts, joints, and dependency structures.
\end{itemize}

\section{Conclusion and Limitations}

\textbf{Conclusion.}
We introduced MechVerse, a benchmark for evaluating image-to-video generation in structured mechanical interaction scenarios. Unlike standard video generation benchmarks that primarily focus on perceptual quality, temporal coherence, and prompt alignment, MechVerse evaluates whether generated videos preserve motion direction, dependency relationships, and interaction dynamics between mechanical components. Our evaluation across open-source and closed-source models shows that current image-to-video systems can often generate visually plausible and temporally smooth videos, but still struggle to produce functionally correct interactions. These results highlight a gap between visual realism and interaction-aware motion generation.

\textbf{Limitations.}
MechVerse focuses on mechanical assemblies and does not cover all forms of physical interaction, such as deformable objects, fluids, human-object interactions, or open-world scenes. Since the benchmark is based on structured assemblies, the results may not fully reflect performance in natural videos. Our evaluation also relies on existing perceptual metrics, instruction-following scores, and human judgments, which may not capture every aspect of physical correctness. Future work can extend MechVerse to broader interaction categories, richer 3D annotations, and more detailed evaluations of force, contact, and long-horizon motion propagation.

\bibliographystyle{plainnat}  
\bibliography{references}

@article{unmesh2023interacting,
  title={Interacting objects: A dataset of object-object interactions for richer dynamic scene representations},
  author={Unmesh, Asim and Jain, Rahul and Shi, Jingyu and Manam, VK Chaithanya and Chi, Hyung-Gun and Chidambaram, Subramanian and Quinn, Alexander and Ramani, Karthik},
  journal={IEEE Robotics and Automation Letters},
  volume={9},
  number={1},
  pages={451--458},
  year={2023},
  publisher={ieee}
}

@inproceedings{ji2020action,
  title={Action genome: Actions as compositions of spatio-temporal scene graphs},
  author={Ji, Jingwei and Krishna, Ranjay and Fei-Fei, Li and Niebles, Juan Carlos},
  booktitle={Proceedings of the IEEE/CVF conference on computer vision and pattern recognition},
  pages={10236--10247},
  year={2020}
}

@inproceedings{miech2019howto100m,
  title     = {{HowTo100M}: Learning a Text-Video Embedding by Watching Hundred Million Narrated Video Clips},
  author    = {Miech, Antoine and Zhukov, Dimitri and Alayrac, Jean-Baptiste and Tapaswi, Makarand and Laptev, Ivan and Sivic, Josef},
  booktitle = {Proceedings of the IEEE/CVF International Conference on Computer Vision (ICCV)},
  year      = {2019}
}

@inproceedings{liu2024evalcrafter,
  title={Evalcrafter: Benchmarking and evaluating large video generation models},
  author={Liu, Yaofang and Cun, Xiaodong and Liu, Xuebo and Wang, Xintao and Zhang, Yong and Chen, Haoxin and Liu, Yang and Zeng, Tieyong and Chan, Raymond and Shan, Ying},
  booktitle={Proceedings of the IEEE/CVF conference on computer vision and pattern recognition},
  pages={22139--22149},
  year={2024}
}

@article{liu2023fetv,
  title={Fetv: A benchmark for fine-grained evaluation of open-domain text-to-video generation},
  author={Liu, Yuanxin and Li, Lei and Ren, Shuhuai and Gao, Rundong and Li, Shicheng and Chen, Sishuo and Sun, Xu and Hou, Lu},
  journal={Advances in Neural Information Processing Systems},
  volume={36},
  pages={62352--62387},
  year={2023}
}

@inproceedings{benshabat2021ikea,
  title     = {The {IKEA} {ASM} Dataset: Understanding People Assembling Furniture through Actions, Objects and Pose},
  author    = {Ben-Shabat, Yizhak and Yu, Xin and Saleh, Fatemeh and Campbell, Dylan and Rodriguez-Opazo, Cristian and Li, Hongdong and Gould, Stephen},
  booktitle = {Proceedings of the IEEE/CVF Winter Conference on Applications of Computer Vision (WACV)},
  year      = {2021}
}

@inproceedings{yang2024unisim,
  title     = {Learning Interactive Real-World Simulators},
  author    = {Yang, Sherry and Du, Yilun and Ghasemipour, Seyed Kamyar Seyed and Tompson, Jonathan and Schuurmans, Dale and Abbeel, Pieter},
  booktitle = {International Conference on Learning Representations (ICLR)},
  year      = {2024}
}

@misc{openai2024sora,
  title        = {Video generation models as world simulators},
  author       = {{OpenAI}},
  year         = {2024},
  howpublished = {\url{https://openai.com/research/video-generation-models-as-world-simulators}},
  note         = {Technical report}
}

@inproceedings{ha2018world,
  title     = {Recurrent World Models Facilitate Policy Evolution},
  author    = {Ha, David and Schmidhuber, J{\"u}rgen},
  booktitle = {Advances in Neural Information Processing Systems (NeurIPS)},
  year      = {2018}
}

@article{meng2024phygenbench,
  title   = {Towards World Simulator: Crafting Physical Commonsense-Based Benchmark for Video Generation},
  author  = {Meng, Fanqing and Liao, Jiaqi and Tan, Xinyu and Shao, Wenqi and Lu, Quanfeng and Zhang, Kaipeng and Cheng, Yu and Qiao, Yu and Luo, Ping},
  journal = {arXiv preprint arXiv:2410.05363},
  year    = {2024}
}

@article{motamed2025physicsiq,
  title   = {Do generative video models understand physical principles?},
  author  = {Motamed, Saman and Culp, Laura and Swersky, Kevin and Jaini, Priyank and Geirhos, Robert},
  journal = {arXiv preprint arXiv:2501.09038},
  year    = {2025}
}

@inproceedings{johnson2017clevr,
  title     = {{CLEVR}: A Diagnostic Dataset for Compositional Language and Elementary Visual Reasoning},
  author    = {Johnson, Justin and Hariharan, Bharath and van der Maaten, Laurens and Fei-Fei, Li and Lawrence Zitnick, C. and Girshick, Ross},
  booktitle = {Proceedings of the IEEE Conference on Computer Vision and Pattern Recognition (CVPR)},
  year      = {2017}
}

@inproceedings{bear2021physion,
  title     = {{Physion}: Evaluating Physical Prediction from Vision in Humans and Machines},
  author    = {Bear, Daniel M. and Wang, Elias and Mrowca, Damian and Binder, Felix J. and Tung, Hsiao-Yu Fish and Pramod, R. T. and Holdaway, Cameron and Tao, Sirui and Smith, Kevin and Sun, Fan-Yun and Fei-Fei, Li and Kanwisher, Nancy and Tenenbaum, Joshua B. and Yamins, Daniel L. K. and Fan, Judith E.},
  booktitle = {Advances in Neural Information Processing Systems (NeurIPS) Datasets and Benchmarks Track},
  year      = {2021}
}

@inproceedings{huang2024vbench,
  title={Vbench: Comprehensive benchmark suite for video generative models},
  author={Huang, Ziqi and He, Yinan and Yu, Jiashuo and Zhang, Fan and Si, Chenyang and Jiang, Yuming and Zhang, Yuanhan and Wu, Tianxing and Jin, Qingyang and Chanpaisit, Nattapol and others},
  booktitle={Proceedings of the IEEE/CVF Conference on Computer Vision and Pattern Recognition},
  pages={21807--21818},
  year={2024}
}

@article{li2025worldmodelbench,
  title={Worldmodelbench: Judging video generation models as world models},
  author={Li, Dacheng and Fang, Yunhao and Chen, Yukang and Yang, Shuo and Cao, Shiyi and Wong, Justin and Luo, Michael and Wang, Xiaolong and Yin, Hongxu and Gonzalez, Joseph E and others},
  journal={arXiv preprint arXiv:2502.20694},
  year={2025}
}

@article{patel2025dynamo,
  title={DYNAMO: Dependency-Aware Deep Learning Framework for Articulated Assembly Motion Prediction},
  author={Patel, Mayank and Jain, Rahul and Unmesh, Asim and Ramani, Karthik},
  journal={arXiv preprint arXiv:2509.12430},
  year={2025}
}

@article{blattmann2023stable,
  title={Stable video diffusion: Scaling latent video diffusion models to large datasets},
  author={Blattmann, Andreas and Dockhorn, Tim and Kulal, Sumith and Mendelevitch, Daniel and Kilian, Maciej and Lorenz, Dominik and Levi, Yam and English, Zion and Voleti, Vikram and Letts, Adam and others},
  journal={arXiv preprint arXiv:2311.15127},
  year={2023}
}

@inproceedings{chen2024videocrafter2,
  title={Videocrafter2: Overcoming data limitations for high-quality video diffusion models},
  author={Chen, Haoxin and Zhang, Yong and Cun, Xiaodong and Xia, Menghan and Wang, Xintao and Weng, Chao and Shan, Ying},
  booktitle={Proceedings of the IEEE/CVF Conference on Computer Vision and Pattern Recognition},
  pages={7310--7320},
  year={2024}
}

@article{ren2024consisti2v,
  title={Consisti2v: Enhancing visual consistency for image-to-video generation},
  author={Ren, Weiming and Yang, Huan and Zhang, Ge and Wei, Cong and Du, Xinrun and Huang, Wenhao and Chen, Wenhu},
  journal={arXiv preprint arXiv:2402.04324},
  year={2024}
}

@inproceedings{girdhar2024factorizing,
  title={Factorizing text-to-video generation by explicit image conditioning},
  author={Girdhar, Rohit and Singh, Mannat and Brown, Andrew and Duval, Quentin and Azadi, Samaneh and Rambhatla, Sai Saketh and Shah, Akbar and Yin, Xi and Parikh, Devi and Misra, Ishan},
  booktitle={European Conference on Computer Vision},
  pages={205--224},
  year={2024},
  organization={Springer}
}

@article{guo2023animatediff,
  title={Animatediff: Animate your personalized text-to-image diffusion models without specific tuning},
  author={Guo, Yuwei and Yang, Ceyuan and Rao, Anyi and Liang, Zhengyang and Wang, Yaohui and Qiao, Yu and Agrawala, Maneesh and Lin, Dahua and Dai, Bo},
  journal={arXiv preprint arXiv:2307.04725},
  year={2023}
}

@article{wang2023modelscope,
  title={Modelscope text-to-video technical report},
  author={Wang, Jiuniu and Yuan, Hangjie and Chen, Dayou and Zhang, Yingya and Wang, Xiang and Zhang, Shiwei},
  journal={arXiv preprint arXiv:2308.06571},
  year={2023}
}

@inproceedings{peebles2023scalable,
  title={Scalable diffusion models with transformers},
  author={Peebles, William and Xie, Saining},
  booktitle={Proceedings of the IEEE/CVF international conference on computer vision},
  pages={4195--4205},
  year={2023}
}

@misc{openai_sora_2025_web,
  author       = {{OpenAI}},
  title        = {Sora},
  year         = {2025},
  howpublished = {\url{https://openai.com/sora/}},
  note         = {Accessed: 2025-10-07}
}

@article{kong2024hunyuanvideo,
  title={Hunyuanvideo: A systematic framework for large video generative models},
  author={Kong, Weijie and Tian, Qi and Zhang, Zijian and Min, Rox and Dai, Zuozhuo and Zhou, Jin and Xiong, Jiangfeng and Li, Xin and Wu, Bo and Zhang, Jianwei and others},
  journal={arXiv preprint arXiv:2412.03603},
  year={2024}
}

@article{wan2025wan,
  title={Wan: Open and advanced large-scale video generative models},
  author={Wan, Team and Wang, Ang and Ai, Baole and Wen, Bin and Mao, Chaojie and Xie, Chen-Wei and Chen, Di and Yu, Feiwu and Zhao, Haiming and Yang, Jianxiao and others},
  journal={arXiv preprint arXiv:2503.20314},
  year={2025}
}

@article{ma2025step,
  title={Step-video-t2v technical report: The practice, challenges, and future of video foundation model},
  author={Ma, Guoqing and Huang, Haoyang and Yan, Kun and Chen, Liangyu and Duan, Nan and Yin, Shengming and Wan, Changyi and Ming, Ranchen and Song, Xiaoniu and Chen, Xing and others},
  journal={arXiv preprint arXiv:2502.10248},
  year={2025}
}

@article{montanaro2024motioncraft,
  title={Motioncraft: Physics-based zero-shot video generation},
  author={Montanaro, Antonio and Savant Aira, Luca and Aiello, Emanuele and Valsesia, Diego and Magli, Enrico},
  journal={Advances in Neural Information Processing Systems},
  volume={37},
  pages={123155--123181},
  year={2024}
}

@article{bansal2024videophy,
  title={Videophy: Evaluating physical commonsense for video generation},
  author={Bansal, Hritik and Lin, Zongyu and Xie, Tianyi and Zong, Zeshun and Yarom, Michal and Bitton, Yonatan and Jiang, Chenfanfu and Sun, Yizhou and Chang, Kai-Wei and Grover, Aditya},
  journal={arXiv preprint arXiv:2406.03520},
  year={2024}
}

@inproceedings{liu2024physgen,
  title={Physgen: Rigid-body physics-grounded image-to-video generation},
  author={Liu, Shaowei and Ren, Zhongzheng and Gupta, Saurabh and Wang, Shenlong},
  booktitle={European Conference on Computer Vision},
  pages={360--378},
  year={2024},
  organization={Springer}
}

@inproceedings{xue2025phyt2v,
  title={Phyt2v: Llm-guided iterative self-refinement for physics-grounded text-to-video generation},
  author={Xue, Qiyao and Yin, Xiangyu and Yang, Boyuan and Gao, Wei},
  booktitle={Proceedings of the Computer Vision and Pattern Recognition Conference},
  pages={18826--18836},
  year={2025}
}

@article{meng2024towards,
  title={Towards world simulator: Crafting physical commonsense-based benchmark for video generation},
  author={Meng, Fanqing and Liao, Jiaqi and Tan, Xinyu and Shao, Wenqi and Lu, Quanfeng and Zhang, Kaipeng and Cheng, Yu and Li, Dianqi and Qiao, Yu and Luo, Ping},
  journal={arXiv preprint arXiv:2410.05363},
  year={2024}
}

@article{le2023differentiable,
  title={Differentiable physics simulation of dynamics-augmented neural objects},
  author={Le Cleac'h, Simon and Yu, Hong-Xing and Guo, Michelle and Howell, Taylor and Gao, Ruohan and Wu, Jiajun and Manchester, Zachary and Schwager, Mac},
  journal={IEEE Robotics and Automation Letters},
  volume={8},
  number={5},
  pages={2780--2787},
  year={2023},
  publisher={IEEE}
}

@article{iliash2024s2o,
  title={S2o: Static to openable enhancement for articulated 3d objects},
  author={Iliash, Denys and Jiang, Hanxiao and Zhang, Yiming and Savva, Manolis and Chang, Angel X},
  journal={arXiv preprint arXiv:2409.18896},
  year={2024}
}

@inproceedings{xiang2020sapien,
  title={Sapien: A simulated part-based interactive environment},
  author={Xiang, Fanbo and Qin, Yuzhe and Mo, Kaichun and Xia, Yikuan and Zhu, Hao and Liu, Fangchen and Liu, Minghua and Jiang, Hanxiao and Yuan, Yifu and Wang, He and others},
  booktitle={Proceedings of the IEEE/CVF conference on computer vision and pattern recognition},
  pages={11097--11107},
  year={2020}
}

@inproceedings{wang2019shape2motion,
  title={Shape2motion: Joint analysis of motion parts and attributes from 3d shapes},
  author={Wang, Xiaogang and Zhou, Bin and Shi, Yahao and Chen, Xiaowu and Zhao, Qinping and Xu, Kai},
  booktitle={Proceedings of the IEEE/CVF Conference on Computer Vision and Pattern Recognition},
  pages={8876--8884},
  year={2019}
}

@article{yan2020rpm,
  title={RPM-Net: recurrent prediction of motion and parts from point cloud},
  author={Yan, Zihao and Hu, Ruizhen and Yan, Xingguang and Chen, Luanmin and Van Kaick, Oliver and Zhang, Hao and Huang, Hui},
  journal={arXiv preprint arXiv:2006.14865},
  year={2020}
}

@inproceedings{xing2024dynamicrafter,
  title={Dynamicrafter: Animating open-domain images with video diffusion priors},
  author={Xing, Jinbo and Xia, Menghan and Zhang, Yong and Chen, Haoxin and Yu, Wangbo and Liu, Hanyuan and Liu, Gongye and Wang, Xintao and Shan, Ying and Wong, Tien-Tsin},
  booktitle={European Conference on Computer Vision},
  pages={399--417},
  year={2024},
  organization={Springer}
}

@article{yang2024cogvideox,
  title={Cogvideox: Text-to-video diffusion models with an expert transformer},
  author={Yang, Zhuoyi and Teng, Jiayan and Zheng, Wendi and Ding, Ming and Huang, Shiyu and Xu, Jiazheng and Yang, Yuanming and Hong, Wenyi and Zhang, Xiaohan and Feng, Guanyu and others},
  journal={arXiv preprint arXiv:2408.06072},
  year={2024}
}

@inproceedings{geng2023gapartnet,
  title={Gapartnet: Cross-category domain-generalizable object perception and manipulation via generalizable and actionable parts},
  author={Geng, Haoran and Xu, Helin and Zhao, Chengyang and Xu, Chao and Yi, Li and Huang, Siyuan and Wang, He},
  booktitle={Proceedings of the IEEE/CVF conference on computer vision and pattern recognition},
  pages={7081--7091},
  year={2023}
}

@article{zheng2023ha,
  title={Ha-vid: A human assembly video dataset for comprehensive assembly knowledge understanding},
  author={Zheng, Hao and Lee, Regina and Lu, Yuqian},
  journal={Advances in Neural Information Processing Systems},
  volume={36},
  pages={67069--67081},
  year={2023}
}

@article{liu2024ikea,
  title={Ikea manuals at work: 4d grounding of assembly instructions on internet videos},
  author={Liu, Yunong and Eyzaguirre, Cristobal and Li, Manling and Khanna, Shubh and Niebles, Juan Carlos and Ravi, Vineeth and Mishra, Saumitra and Liu, Weiyu and Wu, Jiajun},
  year={2024},
  publisher={Advances in Neural Information Processing Systems (NeurIPS) Datasets and~…}
}

@article{wang2025wisa,
  title={Wisa: World simulator assistant for physics-aware text-to-video generation},
  author={Wang, Jing and Ma, Ao and Cao, Ke and Zheng, Jun and Zhang, Zhanjie and Feng, Jiasong and Liu, Shanyuan and Ma, Yuhang and Cheng, Bo and Leng, Dawei and others},
  journal={arXiv preprint arXiv:2503.08153},
  year={2025}
}

@misc{wan27web,
  title        = {Wan 2.7},
  author       = {{Wan AI}},
  year         = {2026},
  howpublished = {\url{https://wan.video/}},
  note         = {Accessed: 2026-05-07}
}

@misc{kling3web,
  title        = {KlingAI 3.0 Series},
  author       = {{Kling AI}},
  year         = {2026},
  howpublished = {\url{https://kling.ai/}},
  note         = {Accessed: 2026-05-07}
}

@misc{happyhorse10web,
  title        = {HappyHorse 1.0},
  author       = {{HappyHorse}},
  year         = {2026},
  howpublished = {\url{https://happyhorse.app/}},
  note         = {Accessed: 2026-05-07}
}

@article{hacohen2024ltxvideo,
  title   = {{LTX-Video}: Realtime Video Latent Diffusion},
  author  = {HaCohen, Yoav and Chiprut, Nisan and Brazowski, Benny and Shalem, Daniel and Moshe, Dudu and Richardson, Eitan and Levin, Eran and Shiran, Guy and Zabari, Nir and Gordon, Ori and Panet, Poriya and Weissbuch, Sapir and Kulikov, Victor and Bitterman, Yaki and Melumian, Zeev and Bibi, Ofir},
  journal = {arXiv preprint arXiv:2501.00103},
  year    = {2024}
}

@article{hacohen2026ltx2,
  title   = {{LTX-2}: Efficient Joint Audio-Visual Foundation Model},
  author  = {HaCohen, Yoav and Brazowski, Benny and Chiprut, Nisan and Bitterman, Yaki and Kvochko, Andrew and Berkowitz, Avishai and Shalem, Daniel and Lifschitz, Daphna and Moshe, Dudu and Porat, Eitan and Richardson, Eitan and Shiran, Guy and Chachy, Itay and Chetboun, Jonathan and Finkelson, Michael and Kupchick, Michael and Zabari, Nir and Guetta, Nitzan and Kotler, Noa and Bibi, Ofir and Gordon, Ori and Panet, Poriya and Benita, Roi and Armon, Shahar and Kulikov, Victor and Inger, Yaron and Shiftan, Yonatan and Melumian, Zeev and Farbman, Zeev},
  journal = {arXiv preprint arXiv:2601.03233},
  year    = {2026}
}


\appendix
\section{MechVerse: Extended Dataset Details}
\label{app:dataset}

\subsection{Dataset Statistics}
Table~\ref{tab:dataset_stats} provides a full breakdown of MechVerse clip counts by category, complexity tier, speed, camera, and motion direction across the full dataset, training split, and test split.

\begin{table}[h]
\caption{MechVerse dataset statistics across splits.}
\label{tab:dataset_stats}
\centering
\begin{tabular}{lrrr}
\toprule
& \textbf{Full} & \textbf{Train} & \textbf{Test} \\
\midrule
Total clips & 21,156 & 18,672 & 1,004 \\
Total assemblies & 1,357 & 1,153 & 204 \\
\midrule
\multicolumn{4}{l}{\textit{Source}} \\
PartNet-Mobility & 15,720 & 14,292 & 476 \\
CAD (MechVerse) & 5,436 & 4,380 & 528 \\
\midrule
\multicolumn{4}{l}{\textit{Complexity}} \\
Easy & 15,720 & 14,292 & 476 \\
Medium & 2,412 & 1,824 & 294 \\
Hard & 3,024 & 2,556 & 234 \\
\midrule
\multicolumn{4}{l}{\textit{Speed}} \\
Slow & 5,742 & 5,033 & 295 \\
Mid & 9,672 & 8,606 & 414 \\
Fast & 5,742 & 5,033 & 295 \\
\midrule
\multicolumn{4}{l}{\textit{Direction}} \\
Forward & 14,508 & 12,909 & 621 \\
Reversed & 6,648 & 5,763 & 383 \\
\midrule
\multicolumn{4}{l}{\textit{Camera}} \\
Cam1 & 7,958 & 6,954 & 1,004 \\
Cam2 & 7,958 & 6,954 & -- \\
Cam3 & 5,240 & 4,764 & -- \\
\bottomrule
\end{tabular}
\end{table}

\subsection{Category List}
MechVerse covers 141 standard categories. Categories sourced from PartNet-Mobility include: Box, Bucket, Dishwasher, Door, Eyeglasses, Fan, Faucet, FoldingChair, Knife, Laptop, Microwave, Oven, Pen, Refrigerator, Safe, StorageFurniture, Table, WashingMachine, and Window. Categories curated as CAD assemblies include: CamAndFollower, CheerleaderToy, EllipticalTrammel, EnginePiston, FourBarLinkage, HobsonJoint, UJoint, and UniqueMechanism. The UniqueMechanism category aggregates 114 individual assembly types for which only one or a small number of instances exist; when expanded, MechVerse covers 141 distinct mechanical categories.

\subsection{Rendering Pipeline}
PartNet-Mobility assemblies were imported into Unity as pre-processed OBJ files with kinematic joint parameters extracted from the dataset's \texttt{mobility\_v2.json} and \texttt{result.json} metadata files. Parts were assigned flat matte colors from a fixed palette using a per-joint material assignment strategy, with one consistent color per joint across all OBJs belonging to that joint, and a randomized palette starting index per assembly to avoid color bias across the dataset. Animation was driven by a frame-accurate normalized time stepping system using \texttt{SetNormalizedTime()}, which steps through exact normalized time positions per frame without relying on Unity's real-time clock, avoiding frame capture stalls observed with \texttt{Time.captureFramerate}. Three cameras were placed at fixed positions in the virtual environment to capture Cam1, Cam2, and Cam3 viewpoints. CAD assemblies were animated using the kinematic solvers built into the respective CAD software, with joint linkages explicitly defined, and rendered from two viewpoints at the same frame specification. All clips are 32 frames at 16 FPS (2 seconds). Reversed clips were generated by reversing the frame order of the corresponding forward clip in post-processing.

\subsection{Annotation Pipeline}\label{app:annotation_pipeline}
\begin{figure}[h]
\centering
\includegraphics[width=\linewidth]{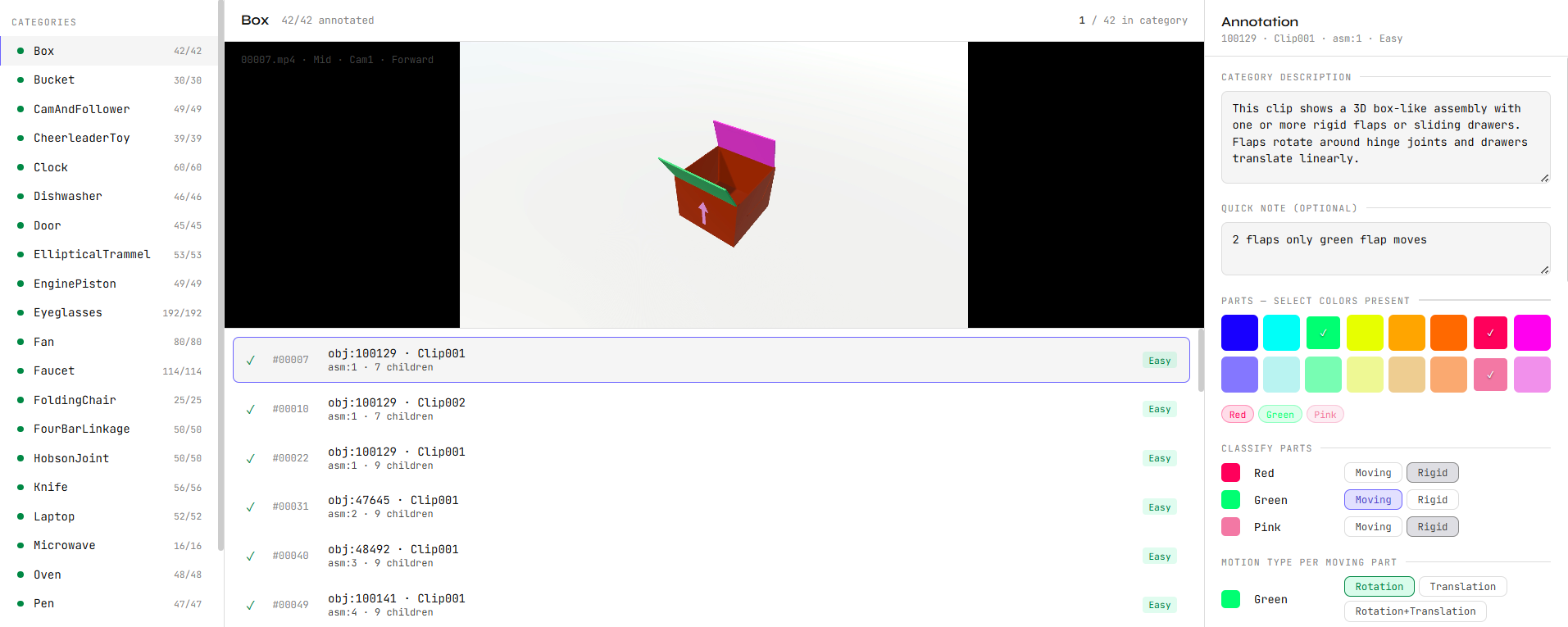}
\caption{Stage 1 annotation web application. Annotators select visible part colors, classify each as moving or stationary, specify motion type and direction per moving part, and review category-level and speed-specific prompt text.}
\label{fig:annotation_webapp}
\end{figure}

\begin{figure}[h]
\centering
\includegraphics[width=\linewidth]{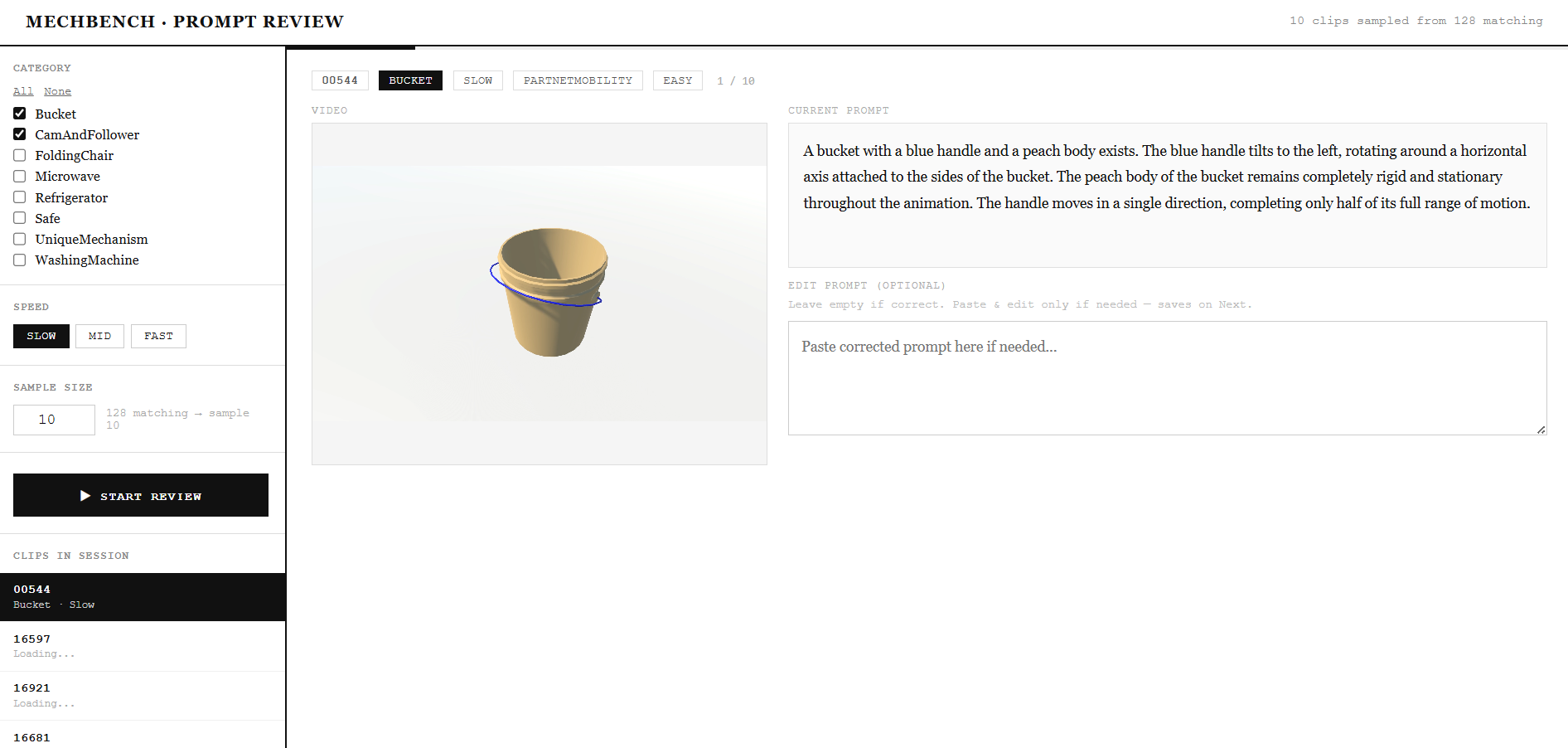}
\caption{Stage 2 prompt review web application. Each clip is displayed alongside its GPT-4o mini generated prompt. Annotators verify accuracy and submit corrections where needed.}
\label{fig:review_webapp}
\end{figure}

Figure~\ref{fig:annotation_webapp} shows the custom web annotation tool used in Stage 1 of the annotation process. For each clip, annotators selected all visible part colors from a fixed palette, classified each color as moving or stationary, selected the motion type for each moving part (Rotation, Translation, Rotation+Translation, Planar), and specified a direction keyword from a controlled vocabulary (Clockwise, Counter-clockwise, Opening, Closing, Extending, Retracting, Folding, Unfolding, Sliding Left, Sliding Right, Sliding Up, Sliding Down, Tilting Left, Tilting Right). A category-level description and an optional quick note capturing assembly-specific details were also recorded. Speed-specific text fields were pre-populated from category-level templates and editable by the annotator. The structured annotation fields were assembled into a draft prompt via a deterministic semantic mapping algorithm and then passed to the GPT-4o mini API for reformatting into fluent prose under a strict system prompt that required all annotated fields to be preserved verbatim and prohibited any addition or omission of content. Figure~\ref{fig:review_webapp} shows the prompt review web application used in Stage 2, where annotators reviewed each clip alongside its generated prompt and submitted corrections where the prompt did not accurately reflect the observed motion.

\subsection{Sample Prompt Structure}
The following shows a representative entry from the final dataset JSON for two speed variants of the same assembly clip, illustrating how prompt content varies across speed while all other annotation fields remain consistent.

\begin{verbatim}
Slow variant:
"A bucket with a blue handle and a pink body has the blue handle tilting to 
the left, rotating around a horizontal axis attached to the sides of the 
bucket. The pink body remains completely rigid and stationary. The handle 
moves in a single direction, completing only half of its full range of motion."

Mid variant:
"A bucket with a blue handle and a pink body features a blue handle that tilts 
to the left, rotating around a horizontal axis attached to the sides of the 
bucket. The pink body remains completely rigid and stationary. The handle 
completes a full motion cycle, sweeping from one side to the other."
\end{verbatim}

\subsection{Preliminary MLLM-Based Annotation Attempts}
\label{app:mllm_annotation}

Prior to developing the expert manual annotation pipeline described in Section~\ref{sec:dataset}, we explored whether multimodal large language models (MLLMs) could be used to automatically annotate clips with structured motion descriptions. We evaluated an MLLM-based pipeline across four objectives: (a) part counting, (b) detection of moving parts, (c) detection of stationary parts, and (d) motion type classification. We further explored three approaches for converting the structured output into natural language prompts suitable for video generation conditioning.

\begin{figure}[t]
\centering
\includegraphics[width=\linewidth]{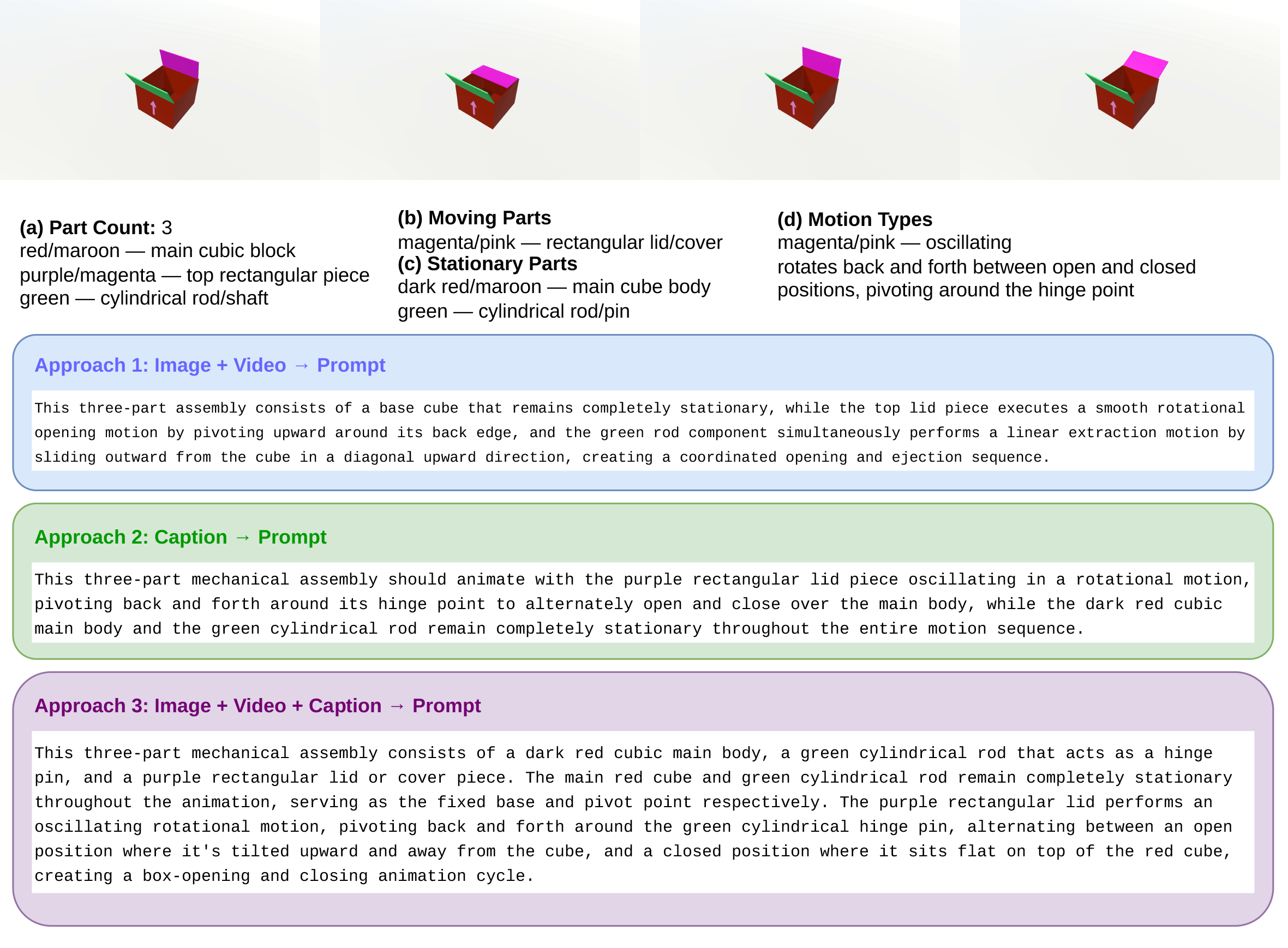}
\caption{MLLM-based annotation results for an Easy-tier assembly (Box). Four temporally-spaced frames are shown alongside structured part analysis (a--d) and generated prompts from all three approaches. The model correctly identifies the oscillating magenta lid, the stationary body and rod, and produces coherent conditioning prompts across all three approaches.}
\label{fig:mllm_easy}
\end{figure}

\noindent\textbf{Part Counting.} A single middle frame from each animation was sent to the MLLM along with a prompt requesting a structured JSON response containing the total part count and per-part color descriptions. This objective worked reliably for simple assemblies with clearly distinguishable part colors.

\noindent\textbf{Moving and Stationary Part Detection.} Three temporally-spaced frames (early, middle, late) were sent together, and the model was asked to compare them and identify which parts moved and which remained stationary. For simple assemblies, this produced broadly correct outputs.

\noindent\textbf{Motion Type Classification.} Motion type classification (rotation, translation, or oscillating) was performed concurrently with part detection. The model returned a motion type field for each identified moving part along with a free-text description of the observed motion.

\noindent\textbf{Prompt Generation Approaches.} Beyond structured annotation, we explored three strategies for generating natural language conditioning prompts:

\begin{description}
    \item[\textbf{Approach 1: Image + Video $\rightarrow$ Prompt}] Early and late frames were provided and the model was prompted to generate a conditioning prompt describing the observed motion.
    \item[\textbf{Approach 2: Caption $\rightarrow$ Prompt}] Only the structured annotation fields (part count, moving parts, stationary parts, motion types) were provided as text, no images. The model generated a prompt purely from textual annotations.
    \item[\textbf{Approach 3: Image + Video + Caption $\rightarrow$ Prompt}] Both the frame pair and structured annotations were provided together, combining visual and textual information.
\end{description}

\begin{figure}[t]
\centering
\includegraphics[width=\linewidth]{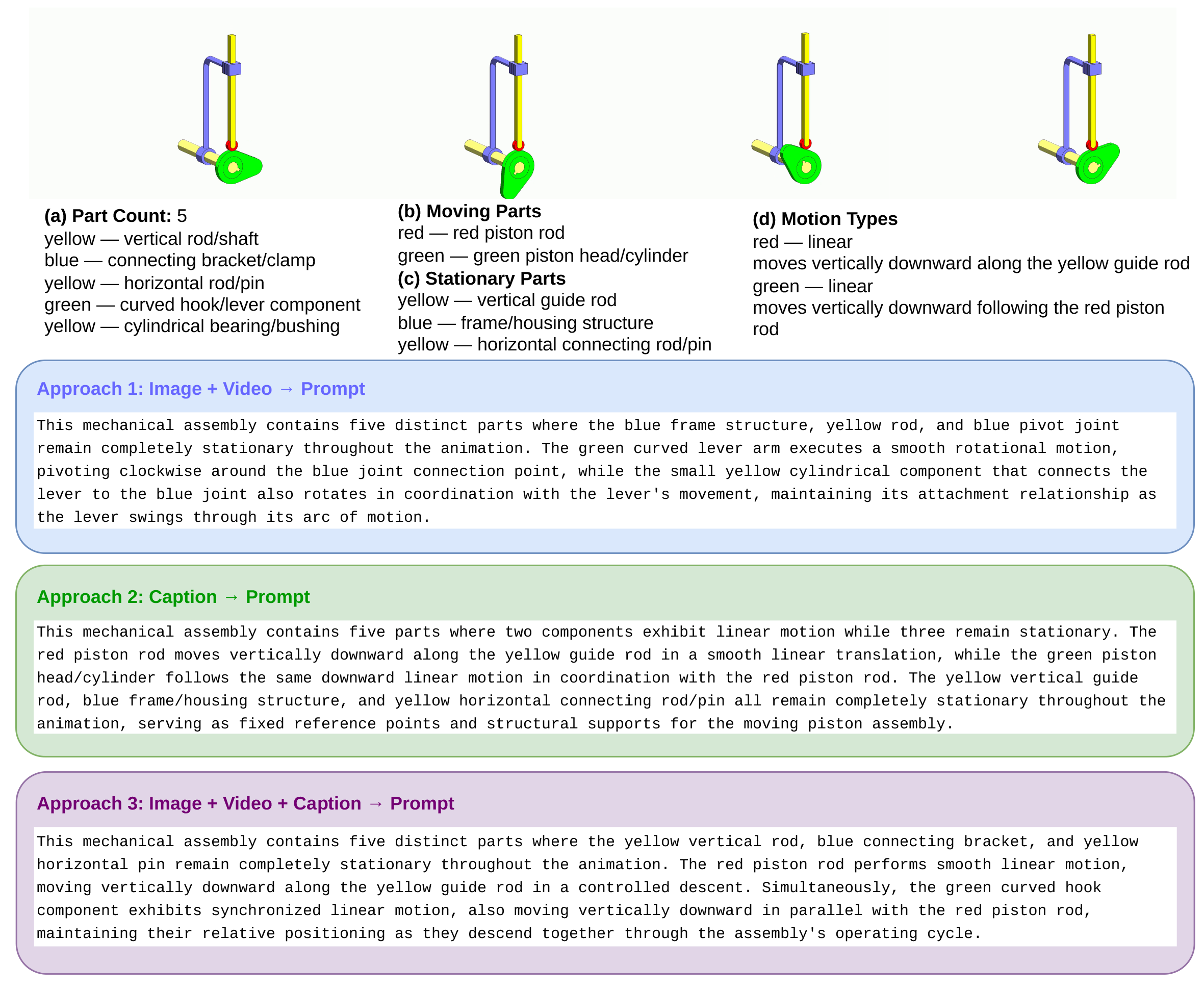}
\caption{MLLM-based annotation failure on a Medium-tier assembly (CamAndFollower). The green cam component is misclassified as exhibiting linear motion (column d) across all three prompt generation approaches, when it is in fact rotating continuously under kinematic coupling. This systematic failure on coupled mechanisms was observed consistently across Medium and Hard tier assemblies and motivated the expert manual annotation pipeline used in MechVerse.}
\label{fig:mllm_hard}
\end{figure}

Figure~\ref{fig:mllm_easy} shows results for an Easy-tier Box assembly. All three approaches produce broadly correct and coherent prompts, the model correctly identifies the oscillating lid, the stationary body, and the hinge-based rotation. This suggests that for simple assemblies with independent part motion and visually distinct colors, MLLM-based annotation is a viable approach.

However, Figure~\ref{fig:mllm_hard} reveals a systematic failure on a Medium-tier CamAndFollower assembly. The green cam component, which performs continuous rotational motion driven by kinematic coupling, is misclassified as linear motion across all three approaches. This error propagates directly into the generated prompts, which describe the cam as moving vertically downward rather than rotating. We observed this class of failure consistently across Medium and Hard tier assemblies, where coupled and non-obvious motion types were regularly misidentified regardless of which prompt generation approach was used. These findings motivated the expert manual annotation pipeline described in Section~\ref{sec:dataset}, in which two annotators with mechanical engineering backgrounds annotated all clips using a controlled vocabulary and structured web interface, ensuring that motion type labels reflect true kinematic behavior rather than superficial visual appearance.

\subsection{Human Evaluation — Per-Model Radar Plots}
\label{app:human_eval_radar}

\begin{figure*}[t]
    \centering
    \includegraphics[width=\linewidth]{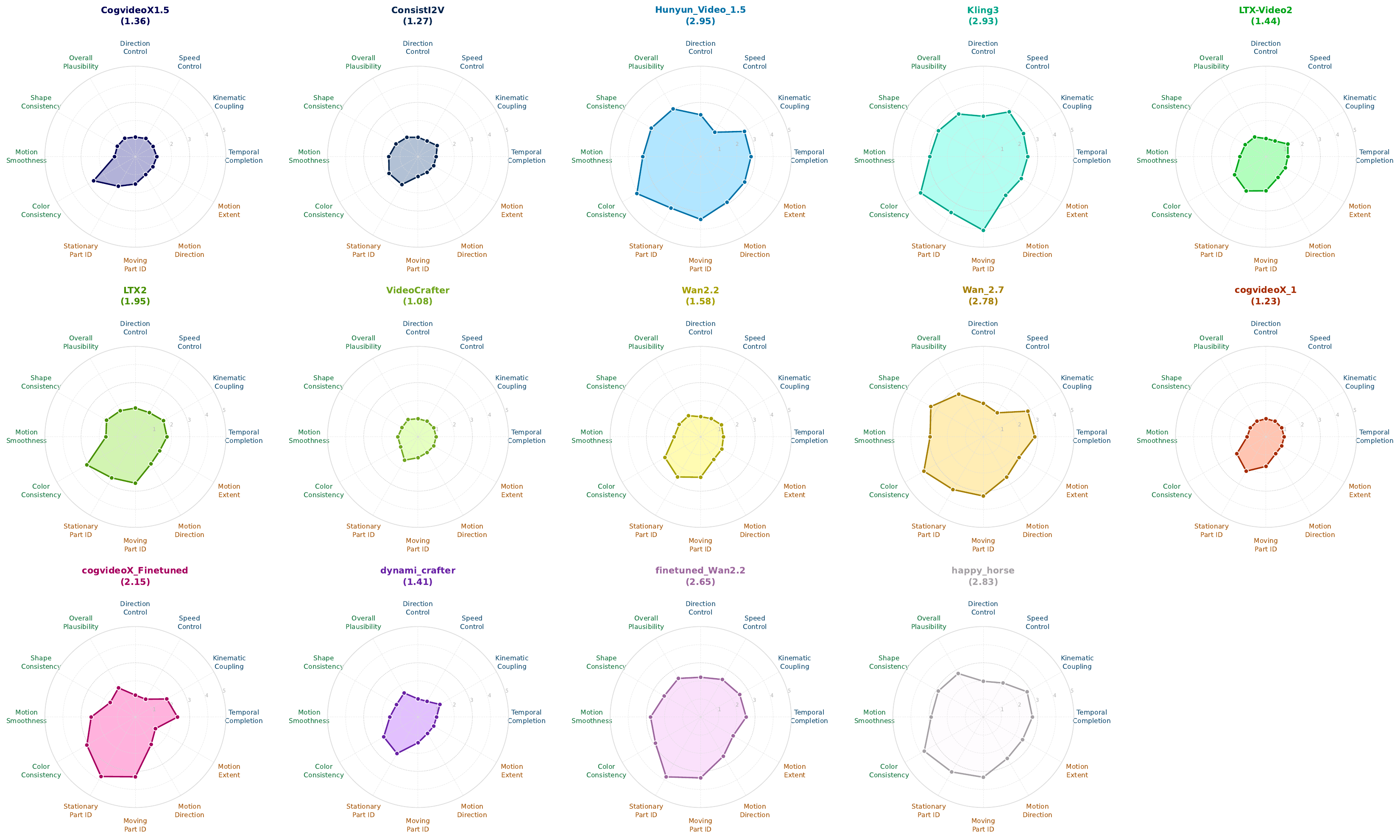}
    \caption{Per-model human evaluation radar plots across all 15 evaluated models. Each subplot shows one model's mean Likert scores (1--5) across 12 evaluation dimensions: Stationary Rigidity (Q1), Correct Part Motion (Q2), Motion Direction (Q3), Motion Extent (Q4), Overall Plausibility (Q5), Shape Consistency (Q6), Motion Smoothness (Q7), Color Consistency (Q8), Kinematic Coupling (Q9), Temporal Completion (Q10), Speed Following (Q11), and Direction Following (Q12). The dotted ring marks the midpoint score of 3.0. Mean score across all dimensions is shown in parentheses below each model name.}
    \label{fig:human_eval_per_model}
\end{figure*}

Figure~\ref{fig:human_eval_per_model} shows the per-model radar plots for all 14 evaluated models across the 12 human evaluation dimensions. Each subplot corresponds to one model, with the mean score across all dimensions shown in the title. The filled area represents the model's score profile; the dotted ring marks the midpoint score of 3.0. Models are sorted by overall mean score. The plots reveal distinct failure signatures across models: most open-source models score consistently low on Kinematic Coupling and Direction Following, while proprietary models such as Kling3 and Horse show broader coverage of the evaluation dimensions. Notably, even the highest-scoring models remain well below the midpoint on several motion-related dimensions, underscoring the difficulty of mechanically-consistent video synthesis.

\subsection{Qualitative Results by Complexity Tier}
\label{app:qualitative}

\begin{figure*}[t]
\centering
\includegraphics[width=\linewidth]{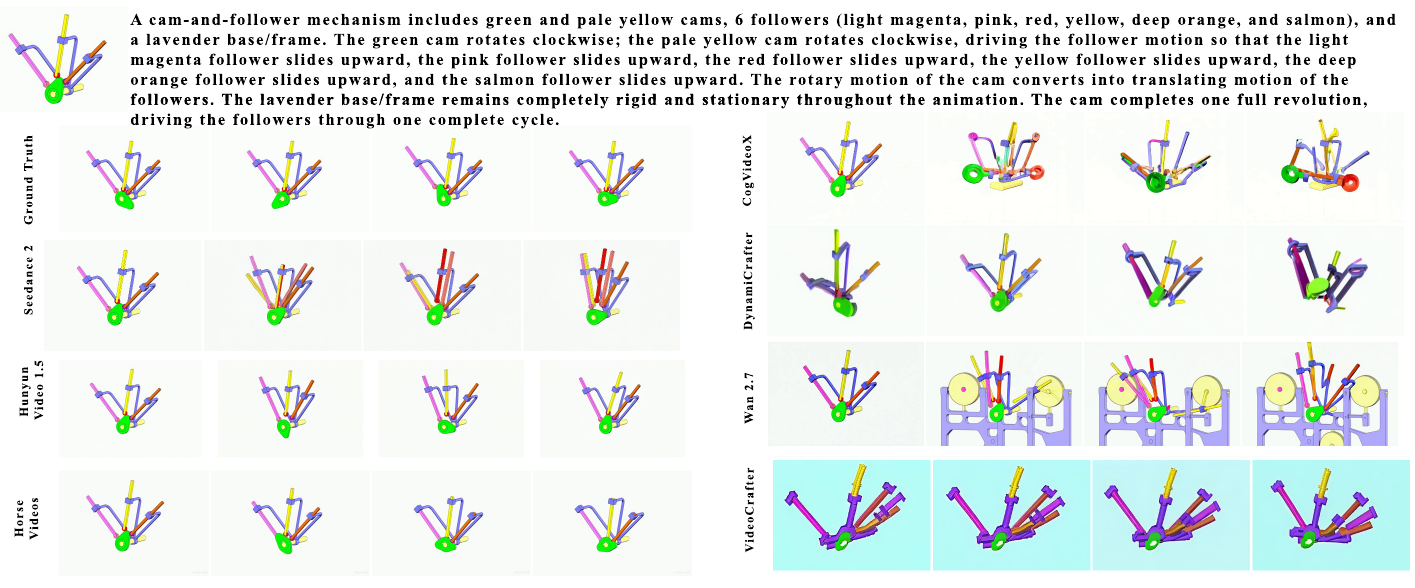}
\caption{Qualitative comparison on a Medium-tier assembly (CamAndFollower). No model correctly reproduces the coupled cam-follower interaction. HunyuanVideo-1.5 produces the closest approximation among open-source models.}
\label{fig:results_mid}
\end{figure*}

\begin{figure*}[t]
\centering
\includegraphics[width=\linewidth]{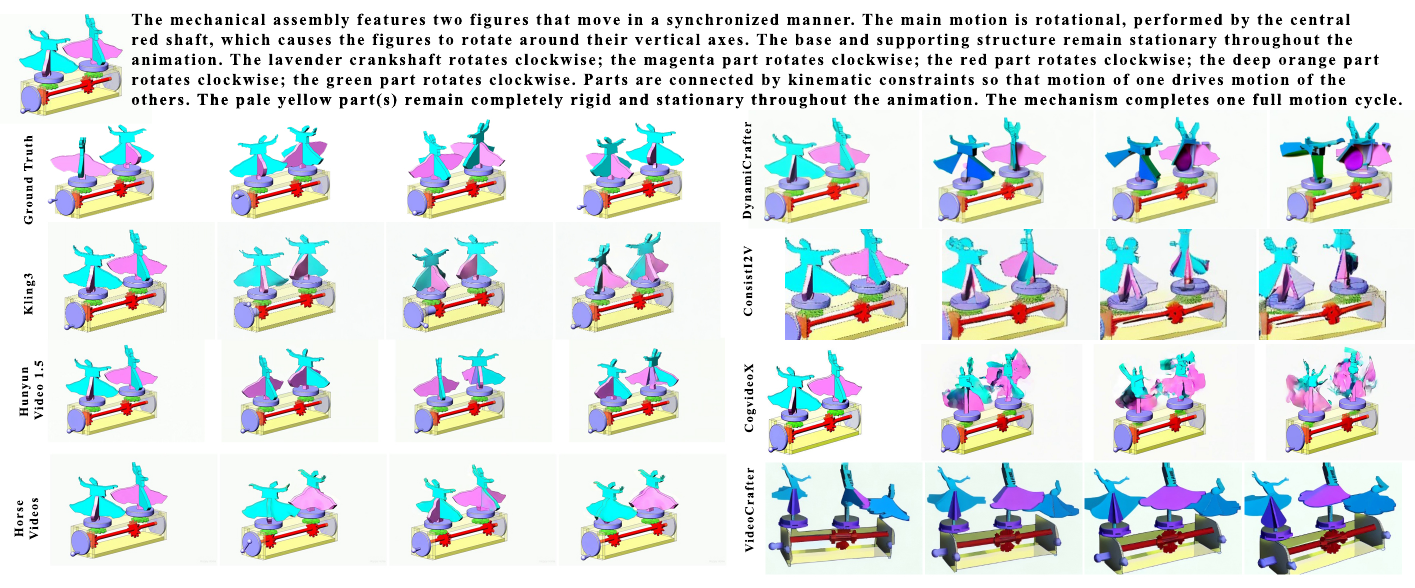}
\caption{Qualitative comparison on a Hard-tier assembly. Top models produce visually plausible animations but fail to propagate motion through the underlying kinematic structure, illustrating why perceptual scores can remain high while mechanism-level correctness is low.}
\label{fig:results_hard}
\end{figure*}

Figures~\ref{fig:results_mid} and~\ref{fig:results_hard} extend the Easy-tier qualitative comparison from the main paper to Medium and Hard assemblies.

On the Medium-tier CamAndFollower assembly (Figure~\ref{fig:results_mid}), HunyuanVideo-1.5 is the only open-source model that produces motion resembling the ground truth. No model correctly generates the coupled cam-follower interaction, while several attempt to rotate the green cam, the followers do not respond with the expected linear translation, violating their kinematic constraints. Kling3 hallucinates a fourth follower, exhibits color flickering, and rotates the followers instead of translating them. Models on the right panel perform uniformly poorly; Wan2.7 hallucinates an entirely different mechanism.

On the Hard-tier assembly (Figure~\ref{fig:results_hard}), the top three models produce visually appealing animations that superficially resemble the ground truth. However, the kinematic constraints in the base mechanism are not respected, the rotating figures move plausibly while the underlying linkages fail to propagate motion correctly. This explains why aggregate human scores can appear relatively high on Hard assemblies (Figure~\ref{fig:human_eval_radar}) even when kinematic coupling scores remain low: models appear to hallucinate motion patterns from visually similar real-world content rather than reasoning from the mechanical structure. Models on the right panel generate poor or incoherent animations with no resemblance to the intended motion.

\end{document}